\documentclass{article}

\pdfoutput=1
\date{}

\usepackage{url}
\usepackage{graphicx}

\usepackage{algorithmic}
\usepackage{algorithm}

\usepackage{multirow}
\usepackage{booktabs}

\usepackage[format=hang, labelformat=parens, font=footnotesize]{subcaption}

\usepackage{authblk}

\begin{document}

\markboth{Maciej Wielgosz, Marcin Pietro\'n}{OpenCL-accelerated object classification in video streams using SP of HTM}

\title{OpenCL-accelerated object classification in video streams using Spatial Pooler of Hierarchical Temporal Memory}

\author{Maciej Wielgosz}
\affil{
AGH University of Science and Technology \\
 Krak\'ow, Poland}

\author{Marcin Pietro\'n}
\affil{Academic Computer Centre CYFRONET,\\
of the University of Science and Technology in Cracow\\
Krak\'ow, Poland}

\maketitle

\begin{abstract}
We present a method to classify objects in video streams using a brain-inspired Hierarchical Temporal Memory (HTM) algorithm. Object classification is a challenging task where humans still significantly outperform machine learning algorithms due to their unique capabilities. We have implemented a system which achieves very promising performance in terms of recognition accuracy. Unfortunately, conducting more advanced experiments is very computationally demanding; some of the trials run on a standard CPU may take as long as several days for 960x540 video streams frames. Therefore we have decided to accelerate selected parts of the system using OpenCL. In particular, we seek to determine to what extent porting selected and computationally demanding parts of a core may speed up calculations. 

The classification accuracy of the system was examined through a series of experiments and the performance was given in terms of F1 score as a function of the number of columns, synapses, $min\_overlap$ and $winners\_set\_size$. The system achieves the highest F1 score of 0.95 and 0.91 for $min\_overlap=4$ and 256 synapses, respectively.  
We have also conduced a series of experiments with different hardware setups and measured CPU/GPU acceleration. The best kernel speed-up of 632x and 207x was reached for 256 synapses and 1024 columns. However, overall acceleration including transfer time was significantly lower and amounted to 6.5x and 3.2x for the same setup.

\end{abstract}

%\keywords{Hierarchical Temporal Memory, OpenCL, GPU, Video processing}

\section{Introduction}
\label{section:intro}

Despite the huge technological growth witnessed nowadays, there are still no autonomous machines available which would be capable of operating in the real world. Such machines would take over most of our tedious everyday duties and clear the way for a breakthrough in Artificial Intelligence. However, such robots need to be able to process inputs in real time, learn, generalize and react to events. This requires building an appropriate processing system which has human--like capabilities.

A mammalian brain is an example of such a system which evolved over millions of years. Despite its apparent complexity there is only one algorithm \cite{Mountcastle} within the brain which governs the body functions. This allows for scalability of the solutions based on the algorithm since more complex systems may be built on a top of the simpler ones just by duplication of the basic structure.

The human brain as a whole has not been completely explored yet, making its artificial implementation and verification a very hard task. However, there are initiatives \cite{humanbrainproject} which have taken up the challenge of simulating and modeling a brain as we know it today. Rather than model the brain, the authors of this paper have adopted a slightly different approach of gradually introducing selected components of Hierarchical Temporal Memory (HTM) to the video processing system with the intention of enhancing its performance. By doing so we aim to develop a complete system\cite{htm} working on the principles of the human brain as they were presented in \cite{Mountcastle, Numenta} with our modification making the algorithm suitable for hardware implementation. Running HTM on CPU is very slow and the algorithm due to its strongly parallel structure is a good candidate for General--Purpose Graphics Processing Unit (GPGPU) and Field--Programmable Gate Array (FPGA) acceleration. Consequently, this paper presents an architecture of GPU implementation of Spatial Pooler (SP). The computationally demanding overlap and inhibition sections of SP were implemented on GPU. 

The rest of the paper is organized as follows. Sections \ref{subsection:HTM} and \ref{subsection:object_detection} provide the background and related work of Hierarchical Temporal Memory and object classification in video streams, respectively. The data flow in the custom--designed system used for the experiments is presented in Section \ref{section:processing_flow} with system architecture described in Section \ref{section:system_description}. Section \ref{section:experiments} provides the results of the experiments. Finally, the conclusions of our research are presented in Section \ref{section:conclusions}.

\subsection{Hierarchical Temporal Memory}
\label{subsection:HTM}

Hierarchical Temporal Memory (HTM) replicates the structural and algorithmic properties of the neocortex. It can be regarded as a memory system which is not programmed, but trained through exposing it to data flow. The process of training is similar to the way humans learn which, in its essence, is about finding latent causes in the acquired content. At the beginning, the HTM has no knowledge of the data stream causes it examines, but through a learning process it explores the causes and captures them in its structure. The training is considered complete when all the latent causes of data are captured and stable. The detailed presentation of HTM is provided in \cite{Numenta, Chen, Rachkovskij}.

HTM constitutes a hierarchy of nodes, where each node performs the same algorithm. The most basic elements (raw and unprocessed data) enter at the bottom of the hierarchy. Each node learns the spatio--temporal pattern of its input and associates it with a given concept. Consequently, each node, no matter where it is in the hierarchy, discovers the causes of its input.
In an HTM, beliefs exist at all levels in the hierarchy and are internal states of each node. They represent probabilities that a cause is active. Each node in an HTM has a fixed number of concepts and a fixed number of output variables. The training process of an HTM starts with a fixed number of possible causes, and in a training process, assigns a meaning to them.

Consequently, the nodes do not increase the number of concepts they cover; instead, over the course of the training, the meaning of the outputs gradually changes. This happens at all levels in the hierarchy simultaneously. Thus the top level of the hierarchy remains with little or no meaning till nodes at the bottom are trained to recognize the basic patterns.

HTM is composed of two main parts, namely Spatial and Temporal Pooler (TP). This paper focuses on Spatial Pooler (SP), aka Pattern Memory, which is employed in the processing flow of the system. It contains columns with synapses connected to the input data \cite{Numenta}. The main role of SP in HTM is finding spatial patterns in the input data. It may be decomposed into three stages:

\begin{itemize}
 \item Overlap calculation (Alg. \ref{alg:overlap}),
 \item Inhibition (Alg. \ref{alg:inhibition}),
 \item Learning.
\end{itemize}

\begin{algorithm}[t]
\caption{Overlap }
\label{alg:overlap}
\begin{algorithmic}[1]
\FORALL{col $\in$ sp.columns} 
\STATE{col.overlap $\leftarrow$ 0} 

\FORALL{syn $\in$ col.connected\_synapses()} 
\STATE{col.overlap $\leftarrow$ col.overlap + syn.active()} 
\ENDFOR

\IF{col.overlap $<$ min\_overlap}
\STATE col.overlap $\leftarrow$ 0
\ELSE
\STATE col.overlap $\leftarrow$ col.overlap * col.boost
\ENDIF
\ENDFOR
\end{algorithmic}
\end{algorithm}

\begin{algorithm}[t]
\caption{Inhibition }
\label{alg:inhibition}
\begin{algorithmic}[1]
\FORALL{col $\in$ sp.columns} 
\STATE{max\_column $\leftarrow$ max(n\_max\_overlap(col, n), 1)} 
\IF{col.overlap $>$ max\_column}
\STATE col.active $\leftarrow$ 1
\ELSE
\STATE col.active $\leftarrow$ 0
\ENDIF
\ENDFOR
\end{algorithmic}
\end{algorithm}

The first two stages are very computationally demanding but can be parallelized. Therefore the authors decided to implement them on GPU in OpenCL. The learning stage, the detailed description of which is provided in the Numenta whitepaper\cite{Numenta}, is implemented on CPU.

The overlap section (Alg. \ref{alg:overlap}) computes $col.overlap$ for every column in SP structure i.e. a number of active and connected synapses. If the number is larger than $col.min\_overlap$, then it is boosted and passed on to the inhibition section (Alg. \ref{alg:inhibition}).

The inhibition stage (Alg. \ref{alg:inhibition}) implements a winner--takes--all procedure where for each column a decision is made as to whether it belongs to a range of $n$ ($winners\_set\_size$) columns of the highest values. The $n\_max\_overlap()$ function performs the comparison.

\subsection{Object classification in video streams}
\label{subsection:object_detection}

\begin{figure}
\centering
\includegraphics[width=0.75\textwidth]{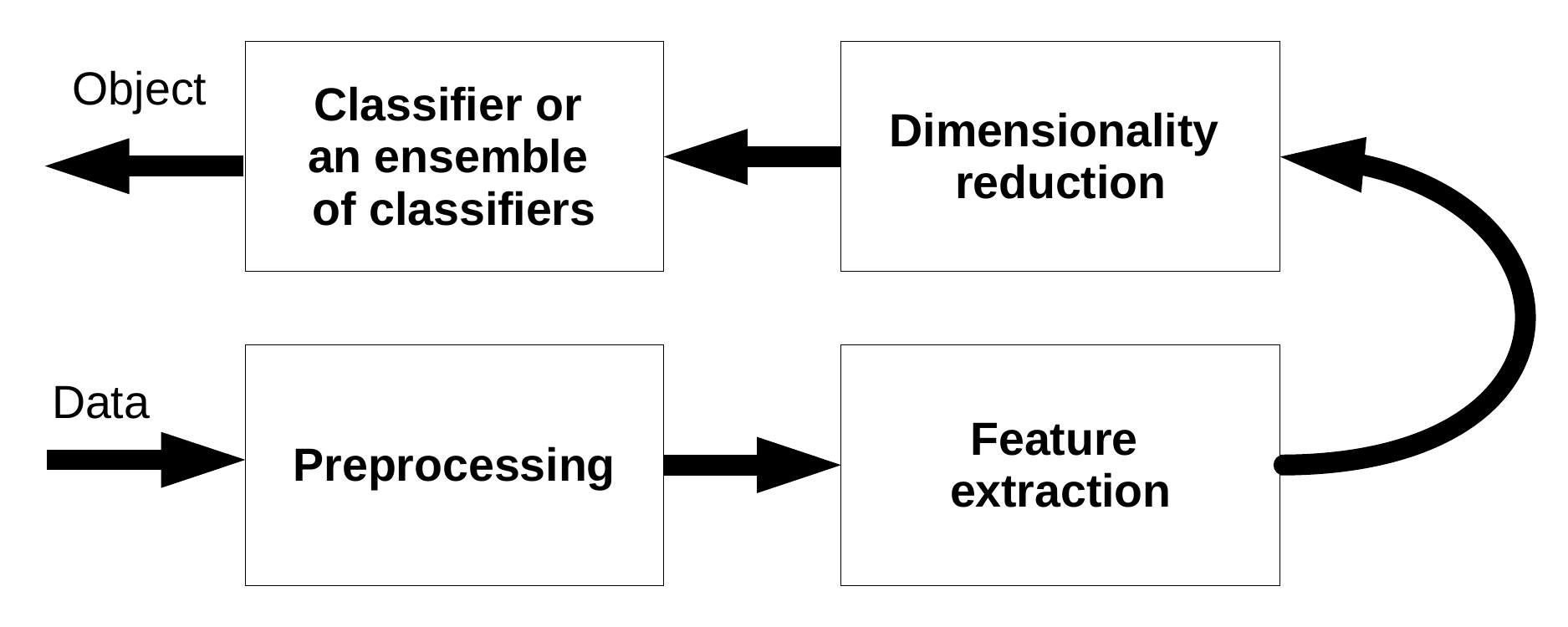}
\caption{Architecture of a video processing system}
\label{fig:architecture_of_a_video_processing_system}
\end{figure}

Most state--of--the--art information extraction systems consist of the following sections: preprocessing, feature extraction, dimensionality reduction and classifier or ensemble of classifiers (Fig. \ref{fig:architecture_of_a_video_processing_system}). Their construction requires expert knowledge as well as familiarity with the data that will be processed \cite{Haibo, Peng}. 

Usually, systems for object classification in video streams are also designed according to this scheme. Consequently, the proper choice of the operations which constitute all the mentioned stages of the system is important and determines the classification result \cite{Lu, Hota, Islam}. One of the most challenging stages is feature extraction, which substantially affects the overall performance of the system.  

There are also systems which take advantage of the spatial--temporal \cite{Numenta} profile  of the data\cite{Castrill,Devarakota,Khan,Bengio}. They are closer to the concept of the solution presented in this paper, which may be considered a hybrid approach since it features components of both schemes.
\section{Processing flow}
\label{section:processing_flow}

The data is fed into the system in a frame--by--frame manner. In the first step, the original frame is turned into a binary image (see \ref{subsection:adaptive_video_encoder}). This conversion constitutes the encoding which allows the generation of input data for the SP processing stage.

Thereafter, the encoded data is fed into the SP. The processing done by the SP effectively maps input to Sparse Distributed Representation (SDR), which then may be passed on to the TP. We do not use TP in this particular application, but the system in general has such a capability. Instead, we substitute TP with histograms to serve a similar purpose.

Histograms of consecutive frames are built from SP output on a per--video basis. The histograms are used as the input data for the SVM classifier which comes next. Classifier maps the results from SDR to the result space (output categories).

\begin{figure}
\centering
\includegraphics[width=0.75\textwidth]{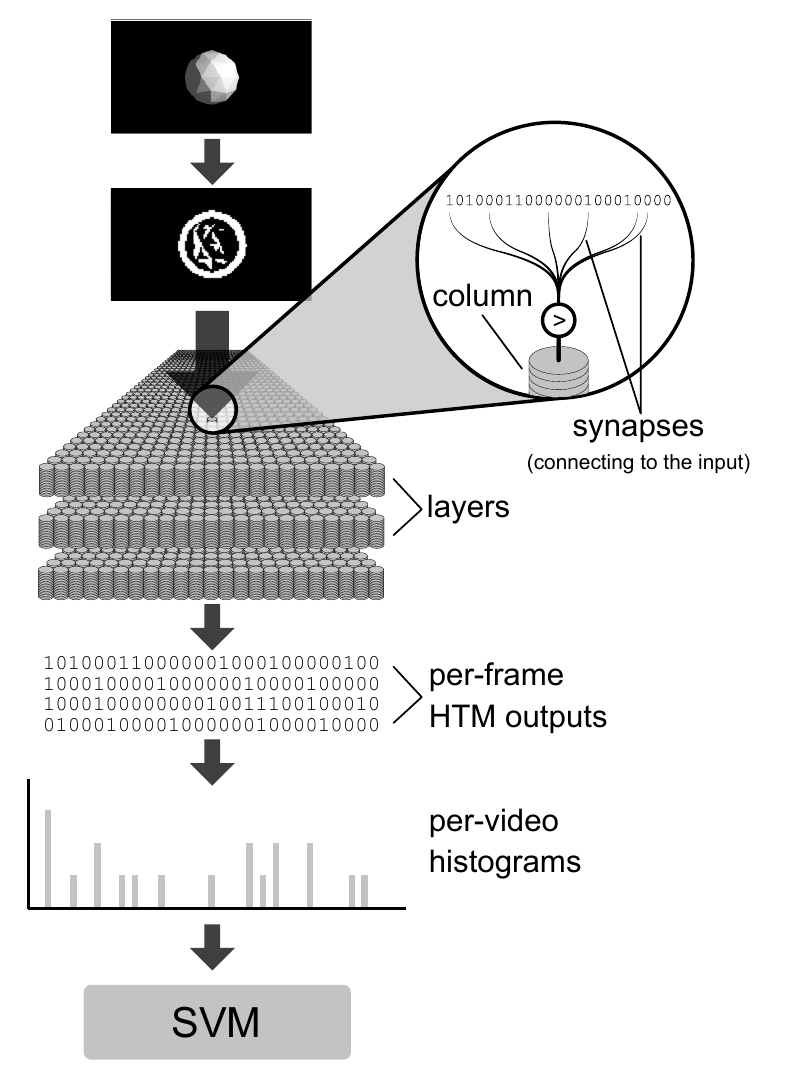}
\caption{Block diagram of the proposed approach}
\label{fig:data_flow}
\end{figure}

The complete processing flow of the system is presented in Fig.~\ref{fig:data_flow}.

\section{System description}
\label{section:system_description}
The system is highly configurable, with numerous parameters responsible for the core HTM's structure, the encoder behavior, statistics rendering, etc. The configuration is stored in a file written in JSON format, which allows it to maintain its readability while providing a clear structure. In addition to the core module, a set of supporting modules has been developed. Most of them are used for feeding video data to the core module, and receiving and analyzing the results. 

\begin{figure}
\centering
    \includegraphics[width=0.85\textwidth]{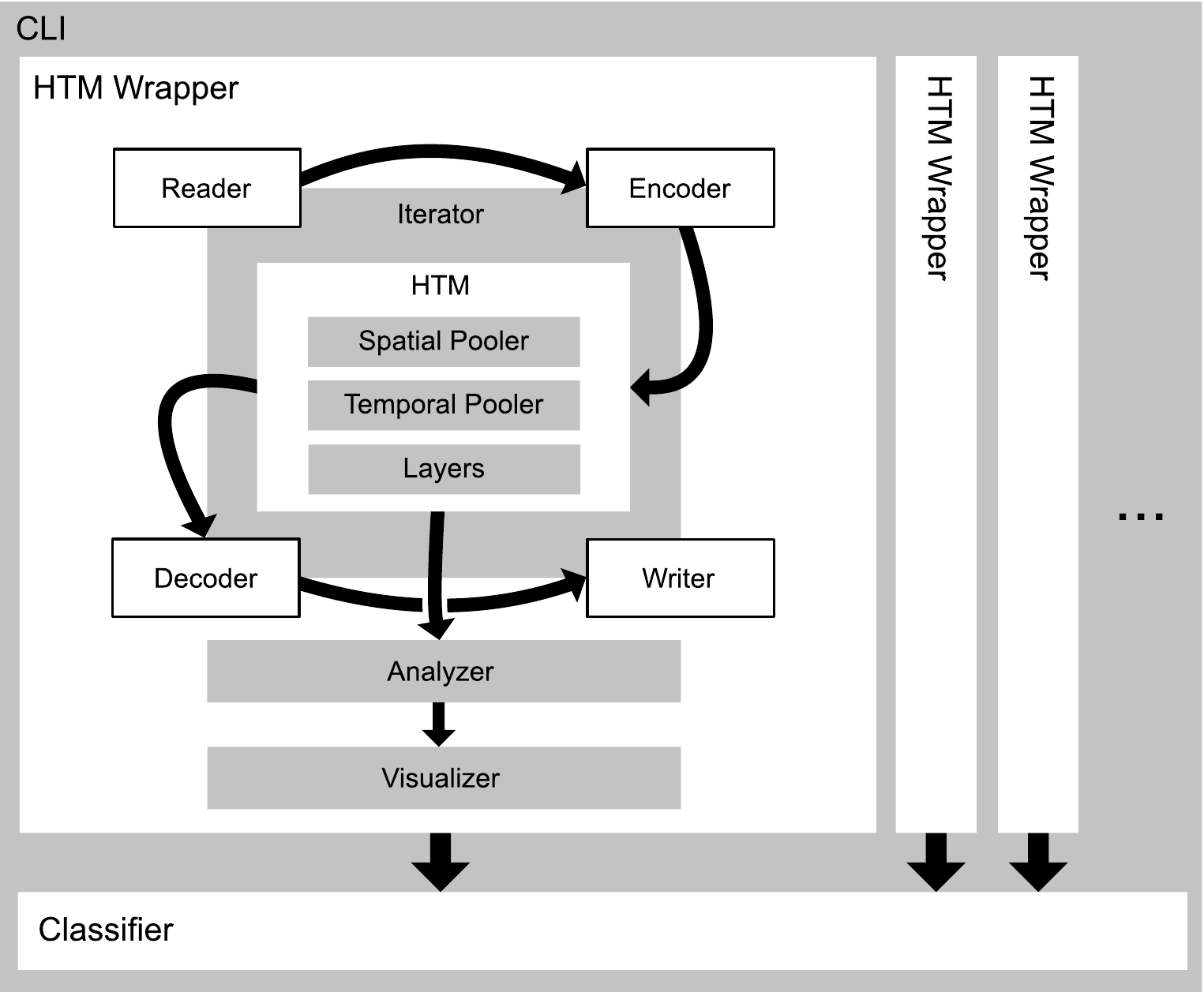}
    \caption{Architecture of the implemented system}
    \label{fig:architecture_of_the_implemented_system}
\end{figure}

The HTM itself is a 'core' module, in addition to the ones necessary for the system to function (responsible for data reading and encoding, as well as results interpretation) and ones created for debugging and statistics gathering purposes. The overall system architecture is depicted in Fig. \ref{fig:architecture_of_the_implemented_system}. The most relevant modules are described in detail below.

\subsection{Outer Structure}
\label{subsection:outer_structure}

The outermost level of system is CLI (Command Line Interface). Depending on the provided command line options, it invokes a particular setup -- either 'Single HTM' or 'Multiple HTMs'. In the 'Single HTM' setup data from all categories is fed into a single HTM instance. 'Multiple HTMs' refers to creating HTM instances on a per--category basis, resulting in an ensemble of one--vs--all detectors.

In both modes the same wrappers encapsulating the actual processing units can be used. A wrapper is created for a particular HTM use -- it is responsible for creating relevant data readers, encoders, decoders and output writers, and for passing them to the iterator -- a part of the core that manages HTM cycles.

After data is processed by the wrapper, the result reaches CLI, which is responsible for further analysis and data presentation -- combining wrappers outputs, gathering statistics, training the classifier used to provide the final results, rendering data visualizations etc. The HTM results are post-processed using a LinearSVM classifier.

\subsubsection{HTM Wrapper}
\label{subsection:htm_wrapper}

As mentioned above, a wrapper is created for a specific use -- the one designed to work with videos will differ from the one tailored for texts. Assembling a wrapper from predefined or newly created modules is the main task of the experiment setup.

The wrapper used in the present system setup creates a reader able to get data from video files and an encoder that converts raw frame data to the required format. The HTM output is neither modified (a pass-through decoder module) nor stored for future reference (a pass-through writer module).

Preparing the processing units to work is not the wrapper's only responsibility -- it also controls the number of executed iterations. The minimum (and default) number of cycles equals a single pass of the learning set, however setups specifying maximum number and/or metrics measuring whether HTM still needs learning are also possible.

The wrapper module also coordinates statistics gathering and visualization on a per-instance basis.

\subsubsection{Adaptive Video Encoder}
\label{subsection:adaptive_video_encoder}

During the encoding process an original video frame is converted to a binary image. Depending on the configuration, the original image can be first reduced in size to trim down the amount of data. After reduction, the color image is converted to a grayscale one, which is later binarized using adaptive thresholding.

Adaptive thresholding uses a potentially different threshold value for each small image region. It gives better results than using a single threshold value for images with varying illumination. In this encoder 'ADAPTIVE\_THRESH\_\-GAUSSIAN\_C' algorithm from OpenCV library\cite{opencv_library} is used -- a threshold value is the weighted sum of neighbourhood values where weights are a gaussian window.

\subsection{HTM Core}

All implemented readers, encoders, decoders and writers provide pre-defined interfaces. Such a solution allows us to separate data acquisition and output storage from the actual processing. The loop consisting of a data retrieval, processing and outputting is executed by the iterator object of the core module.

\subsubsection{HTM}

An HTM object itself consists of a configurable number of layers, a Spatial Pooler and a Temporal Pooler object. Upon each iteration, each layer state is updated by SP and (depending on the configuration) TP, based on the data it receives. In the case of the lowest layer the input is obtained from the encoder, and for the higher ones -- from the previous level. Setting the layer number to zero effectively turns off the HTM, causing the whole module's output to be equal to that of the encoder. This feature was used when comparing performance of 'SVM' only with the 'SP + SVM' ensemble.

Layers consist of columns, which are composed of connectors (containing synapses used in the spatial pooling process) and cells (used in temporal pooling). Cells themselves are built from segments, with each segment containing synapses connecting it to the other cells. This hierarchical structure closely mirrors the one described in the algorithm section.

Every object encapsulates its functionality, making introduction of changes and enhancements trivial, while at the same time providing a clear reference point for modifications. The object-oriented structure also enhances the visibility of a very important HTM feature -- its potential for massive parallelization. One example of that can be a spatial pooling process. The initial system setup used a sequential version of SP. After some tests, a decision to replace it with a concurrent implementation running on a GPU (and an FPGA in the future) was made. The replacement spatial pooler, taking advantage of OpenCL capabilities, was written and plugged into the system without changes to the rest of the architecture.

\subsubsection{Hardware architecture}

\begin{figure}
\centering
    \includegraphics[width=0.85\textwidth]{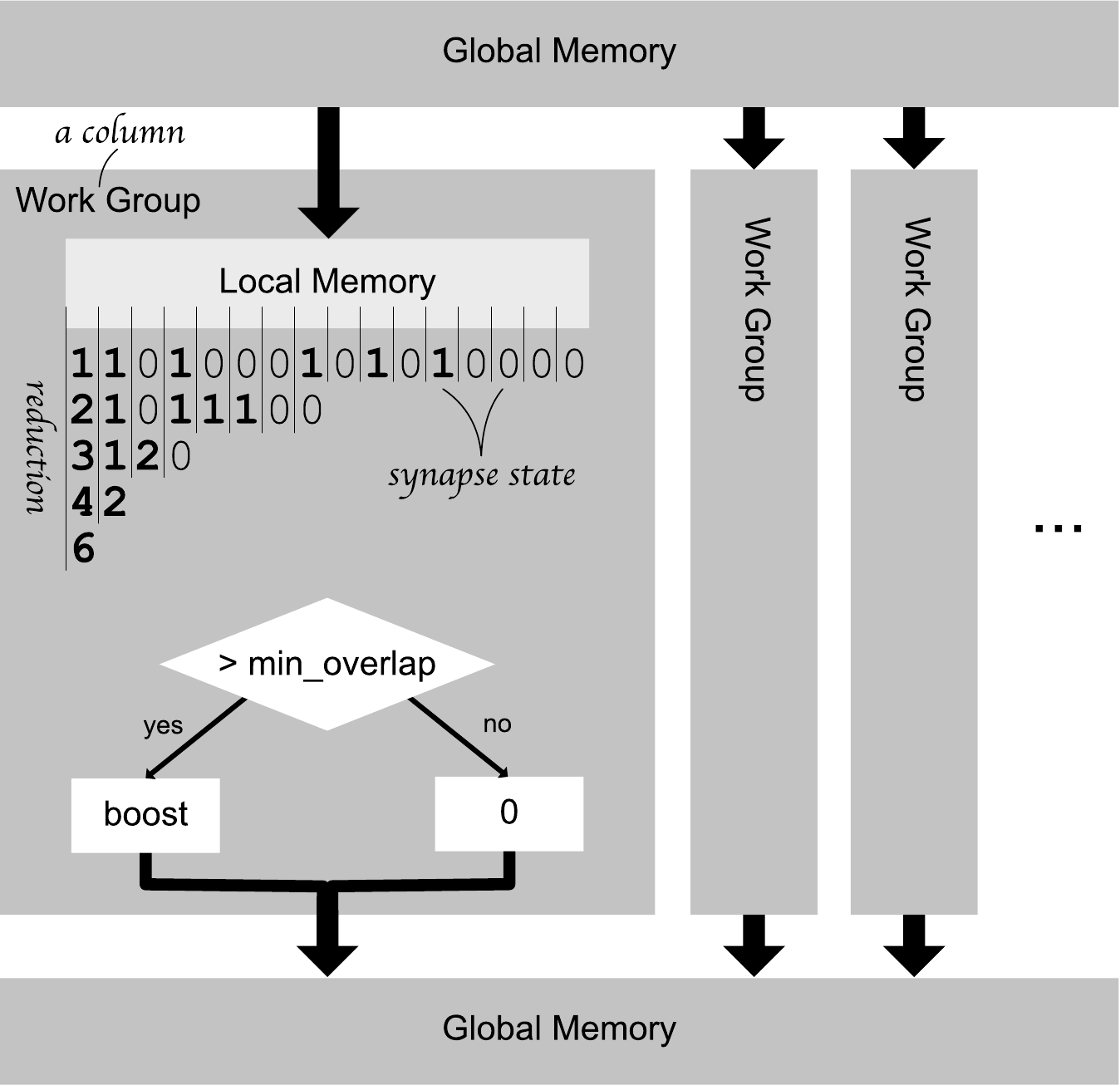}
    \caption{Overlap implemented in OpenCL}
    \label{fig:overlap}
\end{figure}

The overlap calculation is a computationally intensive operation, executed multiple times for every input. Fig. \ref{fig:overlap} presents the hardware architecture of the overlap unit which was implemented in OpenCL. The main idea behind the presented architecture is based on a concept of locating  each column in a separate GPU block (work group). This enables parallel calculation of each column's overlap which is only limited by global--to--local memory data transfer. Once the data is available in the local memory of each work group, a reduction operation is initiated. Intermediate results are stored in the local memory, and in the last stage the results from each block are sent over to the global memory of the GPU. It is worth noting (Fig. \ref{fig:overlap}) that the boost operation\cite{Numenta} is also computed by each kernel within the work group.

\begin{figure}
\centering
    \includegraphics[width=0.75\textwidth]{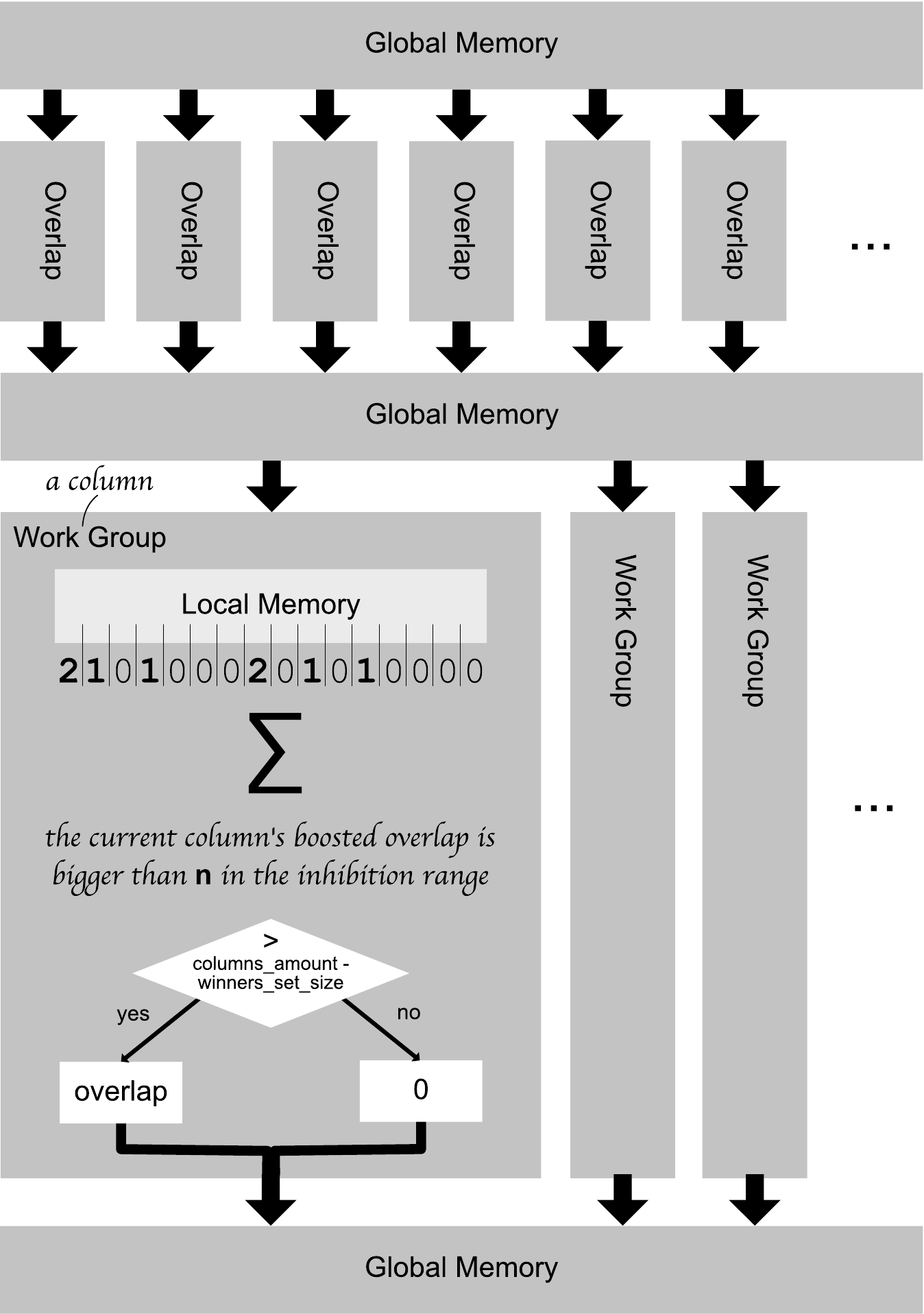}
    \caption{Overlap + Inhibition implemented in OpenCL}
    \label{fig:inhibition}
\end{figure}

The inhibition section presented in Fig. \ref{fig:inhibition} may be considered as an extension of the overlap kernel. It builds up on top of the overlap kernel. The results of the overlap operation are sent back to the global memory of GPU to be fetched again to GPU blocks during the inhibition calculation procedure. The amount of the data required by every work group depends on the inhibition radius. When the overlap data are collected in each work group, a reduction, summation operation and $winners\_set\_size$ comparison is performed. The last operation directly affects the column state by changing it to active or inactive. Extending the overlap module with the logic related to the inhibition calculation improved the performance gain of system as presented in Fig. \ref{fig:pr_overlap_percentage}.

\section{Experiments and the discussion}
\label{section:experiments}

This section presents both quality assessment and acceleration results of the video classification system. It is worth noting that the output of CPU and GPU implementation is not exactly the same due to random initialization of the HTM parameters (e.g. synapses $init\_perm$ values) and learning/testing sets randomization.

All the tests presented in this chapter were performed on Intel(R) Core(TM) i7-4790 CPU @ 3.60GHz with Radeon R9 390 STRIX GPU platform and 32 GB DD3 1600 MHz memory.

\subsection{Experiments setup}

A series of experiments (details of which are provided in Tab.~\ref{tab:experiments_details} and Tab.~\ref{tab:config_params}) was conducted. The experiments allow us to compare the performance of the system featuring Spatial Pooler in the processing flow with the one lacking it, and to measure execution times of both implementations on CPU and GPU.

\begin{table}
\caption{Experiments details}\label{tab:experiments_details}
\begin{tabular}{lll}
\toprule
\multicolumn{2}{l}{Size of a single video frame} & 240x134 \\
\multicolumn{2}{l}{No. of frames in a single video} & 32 \\ 
\multicolumn{2}{l}{Object classes} & cone, cube, cylinder, monkey, sphere, torus \\ 
\multicolumn{2}{l}{No. of classes} & 6 \\ \midrule
\multirow{3}{*}[-0.6em]{Total no. of videos} & all & 6000 \\ \cmidrule{2-3}
 & training & 4800 \\ \cmidrule{2-3}
 & testing & 1200 \\ \midrule
\multirow{3}{*}[-0.6em]{Videos per class} & all & 1000 \\ \cmidrule{2-3}
 & training & 800 \\ \cmidrule{2-3}
 & testing & 200 \\ \midrule
\multirow{3}{*}[-0.6em]{Videos per trial} & all & 100 \\ \cmidrule{2-3}
 & training & 80 \\ \cmidrule{2-3}
 & testing & 20 \\
 \bottomrule
\end{tabular}
\end{table}

\begin{table}
\caption{Basic configuration parameters}\label{tab:config_params}
\centering
\begin{tabular}{lc}
\toprule
No. of columns & 2048\\ 
No. of synapses per column  & 128\\ 
Perm value increment & 0.1\\ 
Perm value decrement & 0.1\\ 
Min overlap & 8\\ 
Winners set size & 40 \\ 
Initial perm value & 0.21 \\ 
Initial inhibition radius & 80 \\ \bottomrule
\end{tabular}
\end{table}

The experiments were conducted using a 'Single HTM' setup (see \ref{subsection:outer_structure}). For each trial, the system was trained in the learning mode with 80\% of available data (80 videos of each class randomly selected from a pool of 800) and then was tested with the remaining 20\% of the data in the testing mode (20 videos per class selected out of 200).

During the course of an experiment the value of a single configuration parameter was changed, while the rest remained as in Tab. \ref{tab:config_params}. Each generated configuration was then used to run tests both on GPU and CPU using OpenCL inhibition kernel. Additionally, the same experiments with columns and synapses were conducted also for the overlap kernel (Fig. \ref{fig:pr_overlap_percentage}).

\subsection{Dataset}

\begin{figure}
\centering
\includegraphics[width=\textwidth]{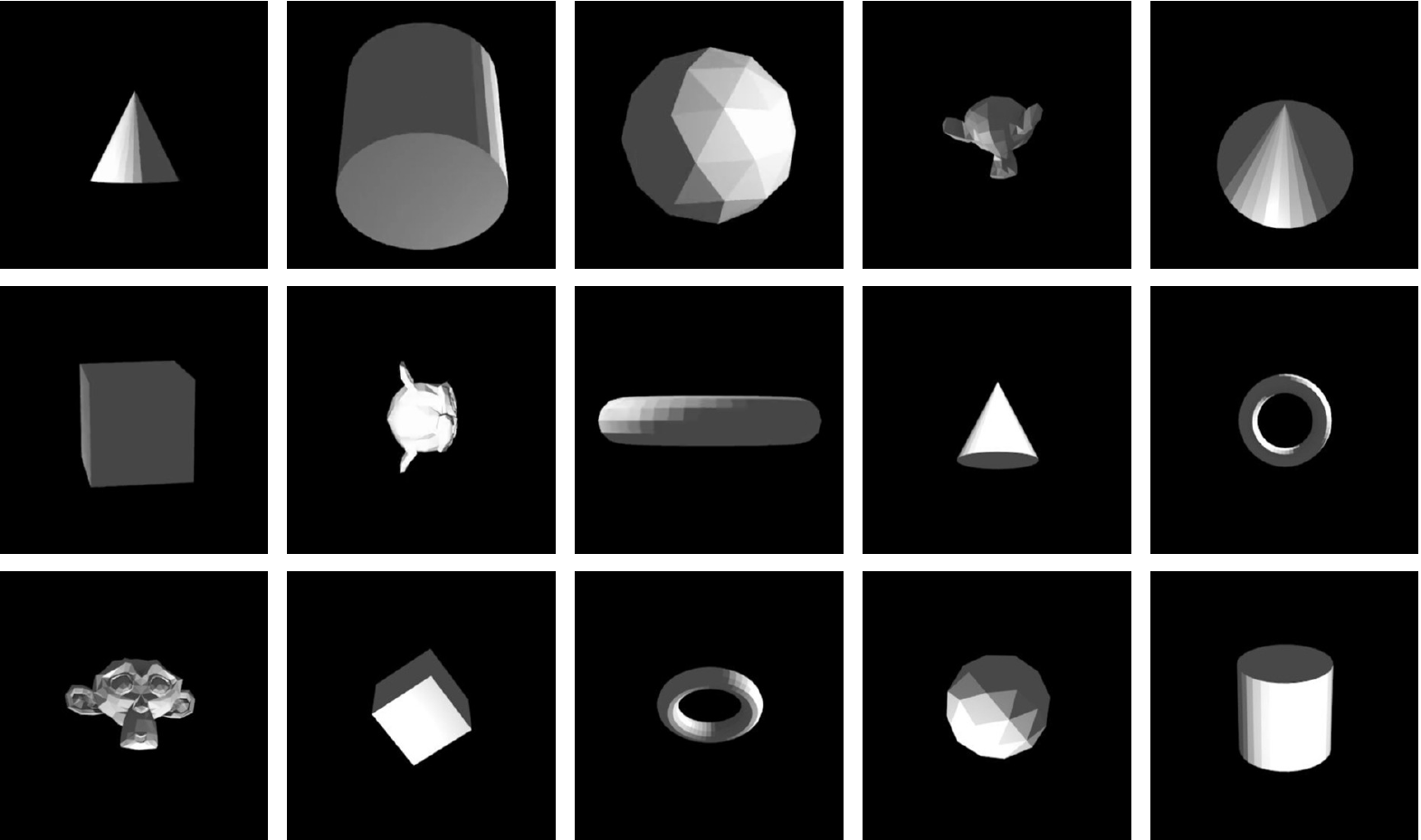}
\caption{Sample frames of different shapes rendered in Blender}
\label{fig:video_screenshots}
\end{figure}

The challenging part involved generation of sample videos for testing. The videos had to meet a series of requirements such as object location, camera location and object--camera distance. Consequently, a dedicated application was used to generate the videos (i.e. Blender \cite{blender}). Original rendered videos had a size of 960x540 pixels and showed a single, centered, stationary object with camera moving around it (Fig. \ref{fig:video_screenshots}).

For the experiments, the dataset (available online\cite{shapes_dataset}) based on the rendered videos was created, with the frame resized to 240x134 pixels. The initial testing showed that reducing the frame size has a very small impact on SVM results (used as a baseline for comparison), while significantly shortening the HTM calculation time.

\subsection{Quality assessment}
\label{subsection:quality_assessment_results}

\begin{figure}
\centering
\begin{subfigure}{0.65\textwidth}
    \includegraphics[width=\textwidth]{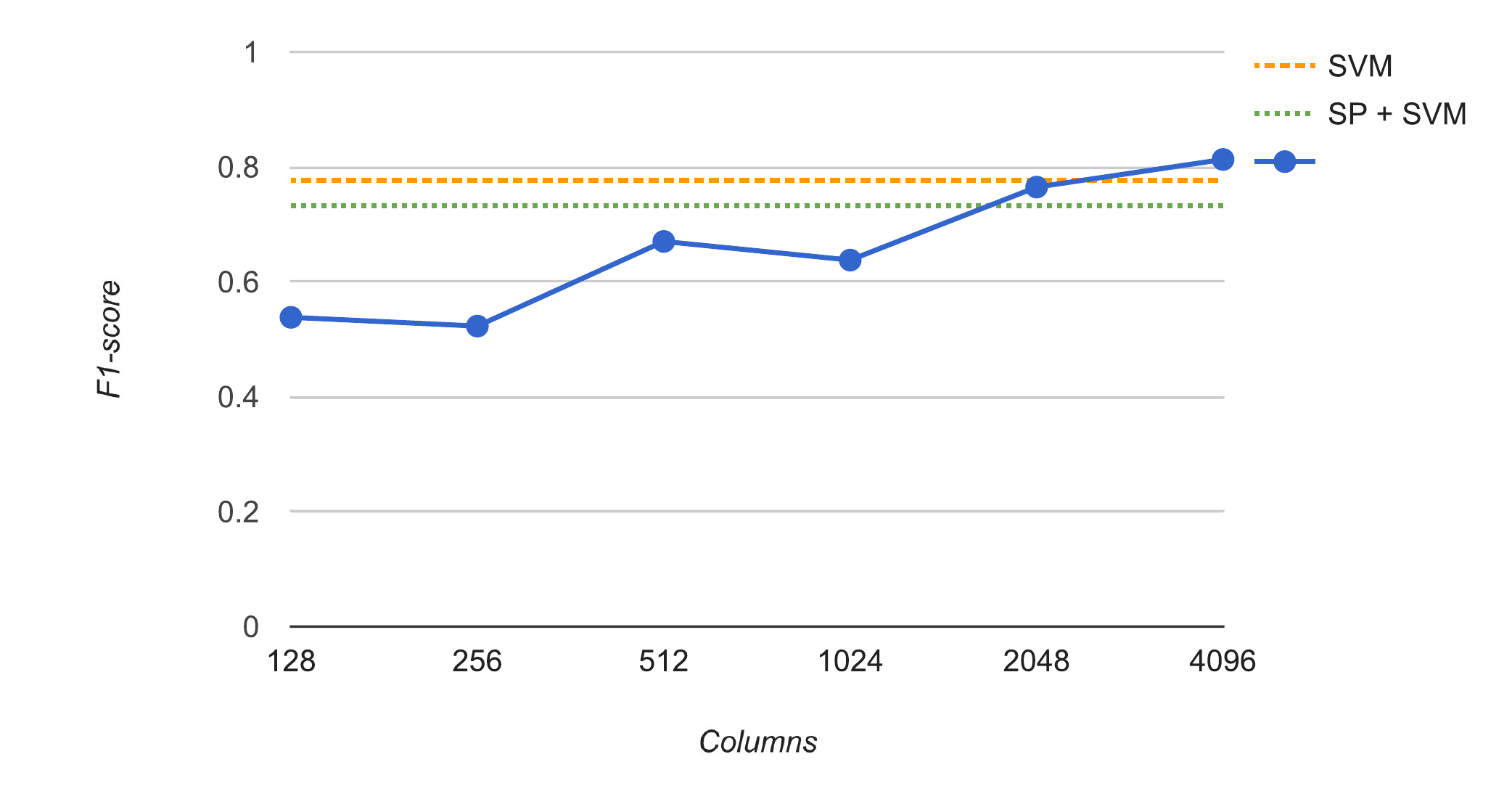}
\end{subfigure}
\begin{subfigure}{0.65\textwidth}
    \includegraphics[width=\textwidth]{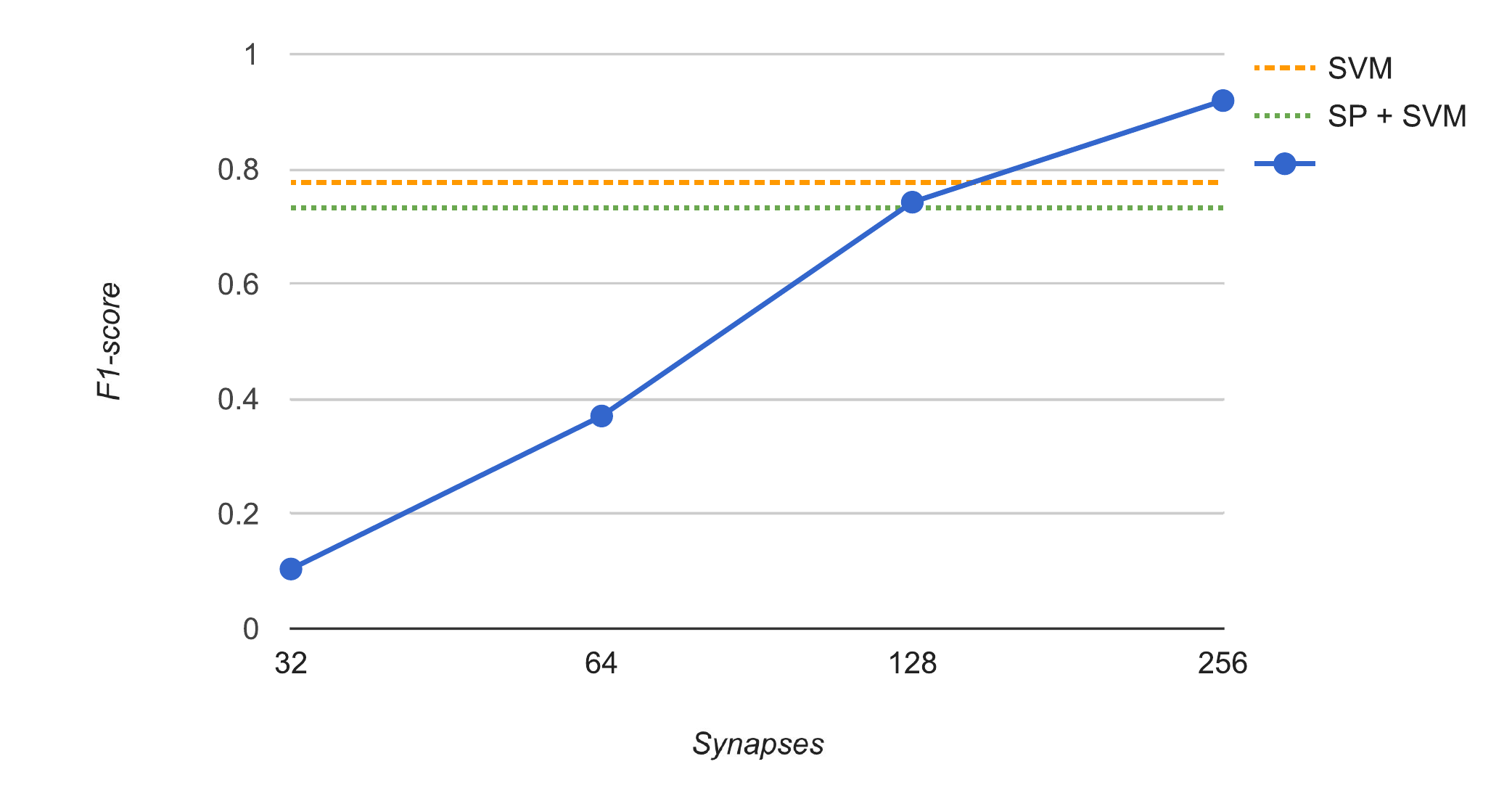}
\end{subfigure}
\begin{subfigure}{0.65\textwidth}
    \includegraphics[width=\textwidth]{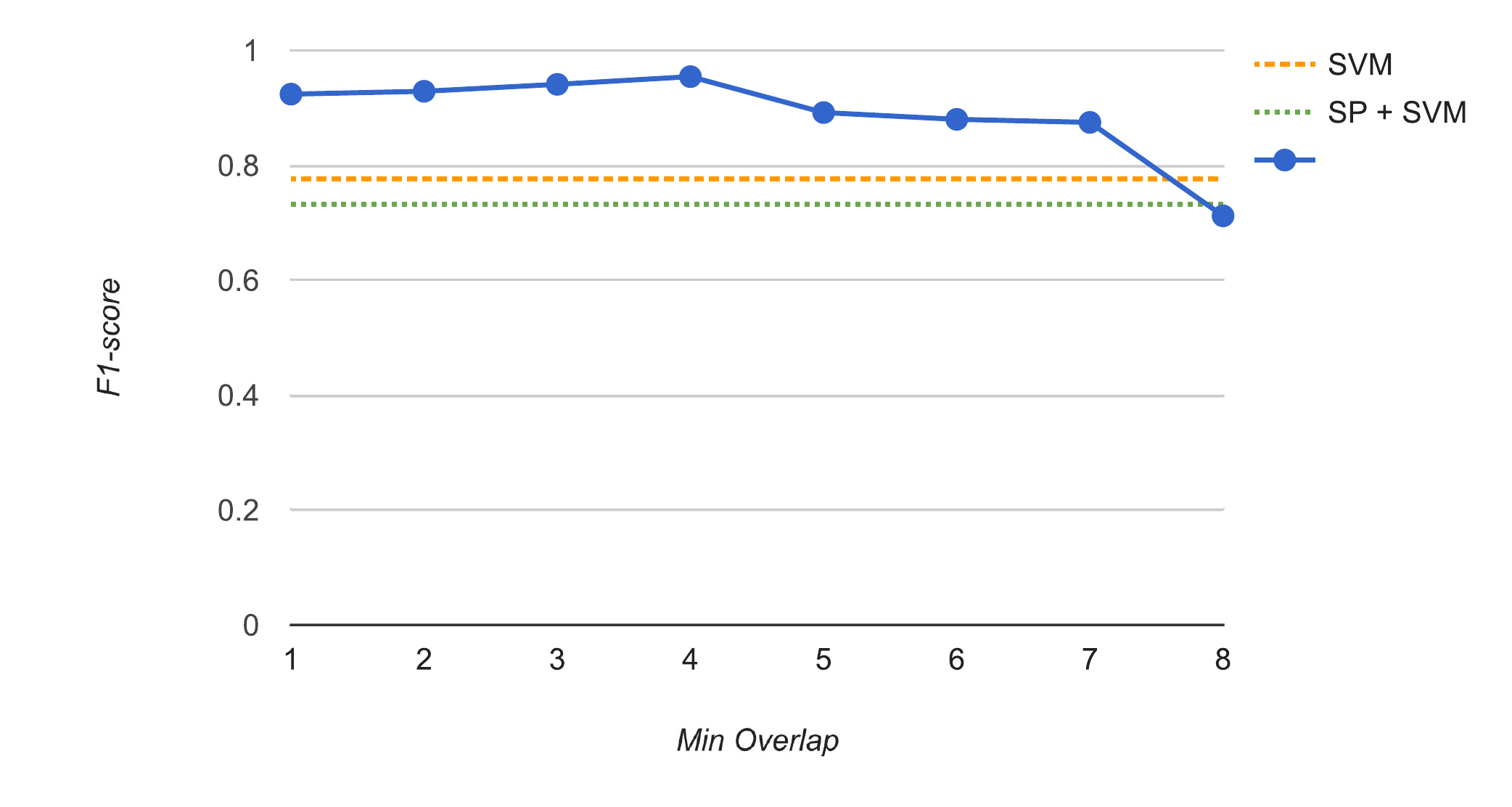}
\end{subfigure}
\begin{subfigure}{0.65\textwidth}
    \includegraphics[width=\textwidth]{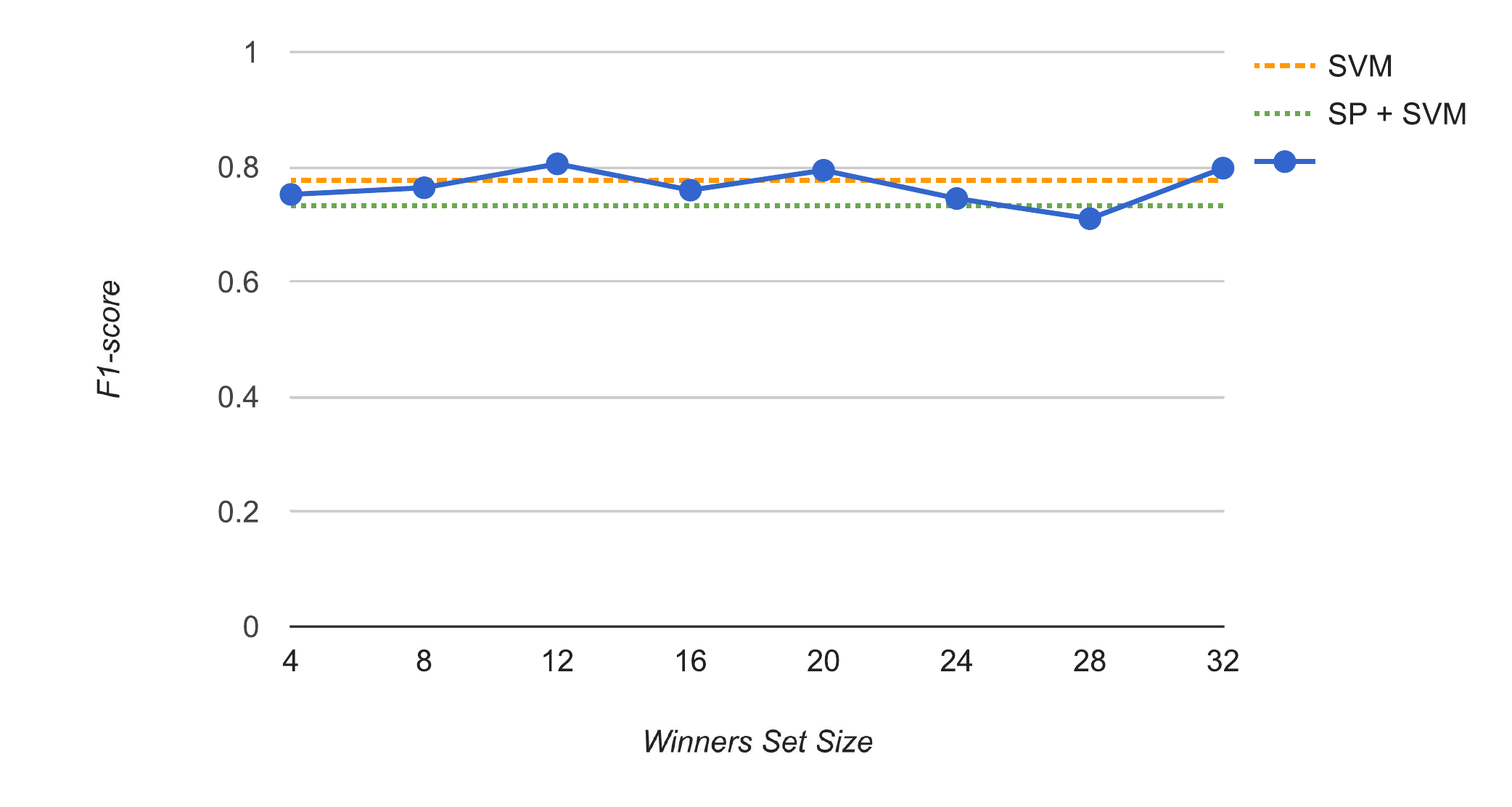}
\end{subfigure}
\caption{Average F1 scores as a function of different SP configuration parameters}
\label{fig:rr_quality_parameters}
\end{figure}

The F1 score is used as a quality evaluation of the experiments' results presented in this paper.  
The precision and recall for corresponding clusters are calculated as follows:
\begin{equation}
 Recall(i, j) = \frac{n_{ij}}{n_i},
 \label{eq:recallClustering}
\end{equation}
\begin{equation}
 Precision(i, j) = \frac{n_{ij}}{n_j},
 \label{eq:precisionClustering}
\end{equation}
where $n_{ij}$ is the number of items of class \textit{i} that are classified as members of cluster \textit{j}, while $n_j$ and $n_i$ are the numbers of items in cluster \textit{j} and class \textit{i}, respectively.
The cluster's F1 score is given by the following formula:
\begin{equation}
 F(i, j) = 2 \cdot \frac{Recall(i, j)Precision(i, j)}{Precision(i, j) + Recall(i,j)}.
 \label{eq:FractionalFmeasure}
\end{equation}
The overall quality of the classification can be obtained by taking the weighted average F1 scores for each class. It is given by the equation:
\begin{equation}
 F1 = \sum\limits_{i} \frac{n_i}{n}max{F(i,j)},
 \label{eq:OverallFmeasure}
\end{equation}
where the maximum is taken over all clusters and \textit{n} is the number of all objects. The F1 score value ranges from 0 to 1, with a higher value indicating a higher clustering quality.

In each experiment presented in Fig. \ref{fig:rr_quality_parameters} one  of the parameters was changed. 'SP + SVM' refers to the baseline results obtained with the proposed system using configuration values from Tab. \ref{tab:config_params}. It is worth noting that despite the superiority of the baseline 'SVM' setup, the 'SP + SVM' performance in selected cases is better than it is for 'SVM'. Especially, the number of synapses and the $min\_overlap$ value affects the performance of the module i.e. a rise in the number of synapses and a drop in the $min\_overlap$ value leads to better classification results. For every value of $winners\_set\_size$ the results remain on the same level with low fluctuation around the baseline. This results from the relationship between the inhibition radius and the $winners\_set\_size$ parameter. Change of the $winners\_set\_size$ is compensated by appropriate adaptation of the inhibition radius \cite{Numenta}.

\subsection{Acceleration results}
\label{subsection:acceleration_results}

A series of comparative tests were carried out for columns, synapses, $min\_overlap$ and $winners\_set\_size$. Two different test types  were conducted, namely \textit{GPU vs CPU OCL} denoted also as \textit{OCL} and \textit{GPU vs CPU kernel} referred to as \textit{kernel} in the text. The first one accounts for the complete execution time of the examined procedures i.e. data preparation, data transfer in both directions and kernel execution\cite{kloeckner_pycuda_2012}. The second test type embraces only kernel execution. 

\begin{figure}[p]
\centering
\begin{subfigure}[t]{0.48\textwidth}
    \includegraphics[width=\textwidth]{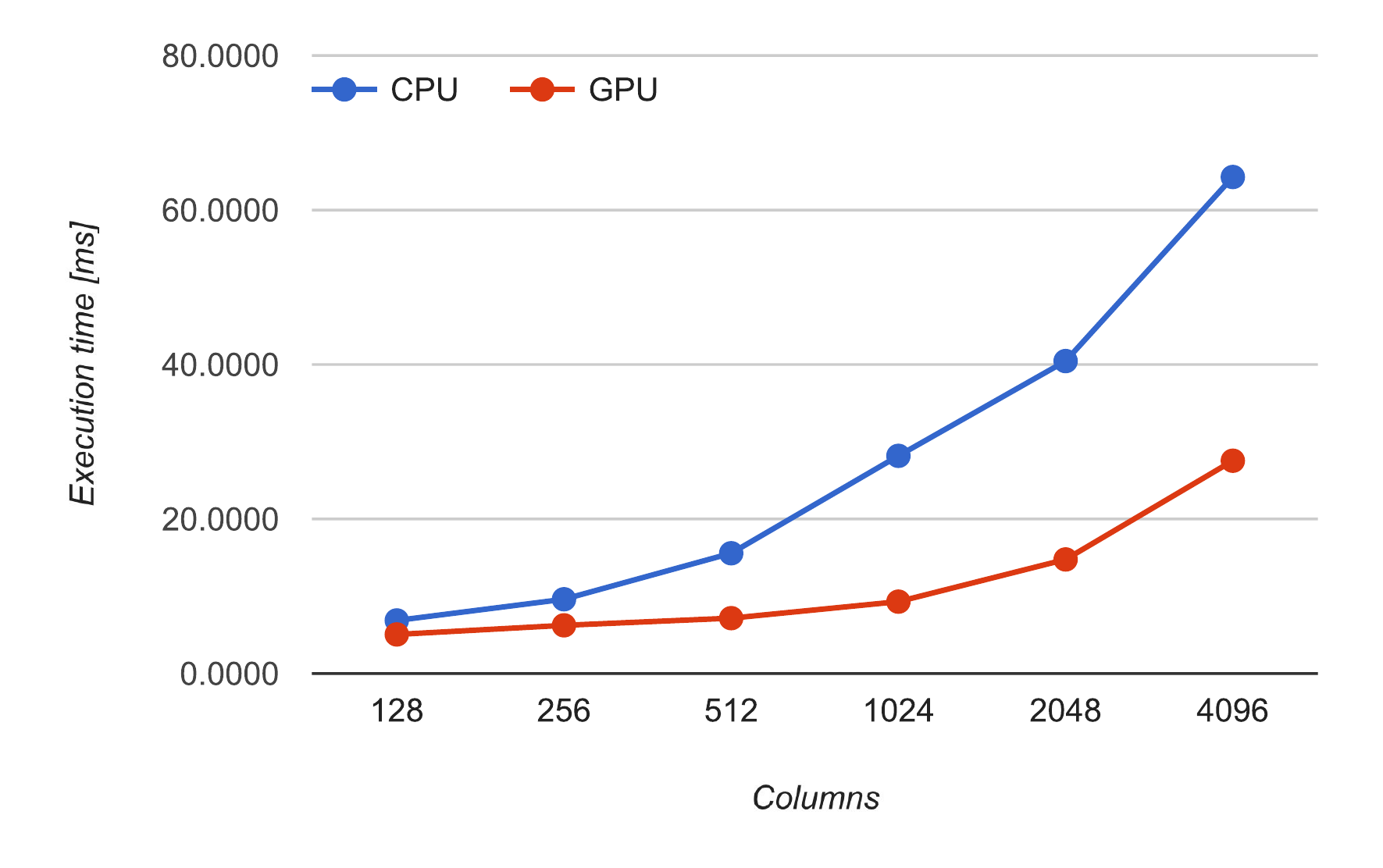}
    \caption{Average OCL kernel exec time}
\end{subfigure}
\begin{subfigure}[t]{0.48\textwidth}
    \includegraphics[width=\textwidth]{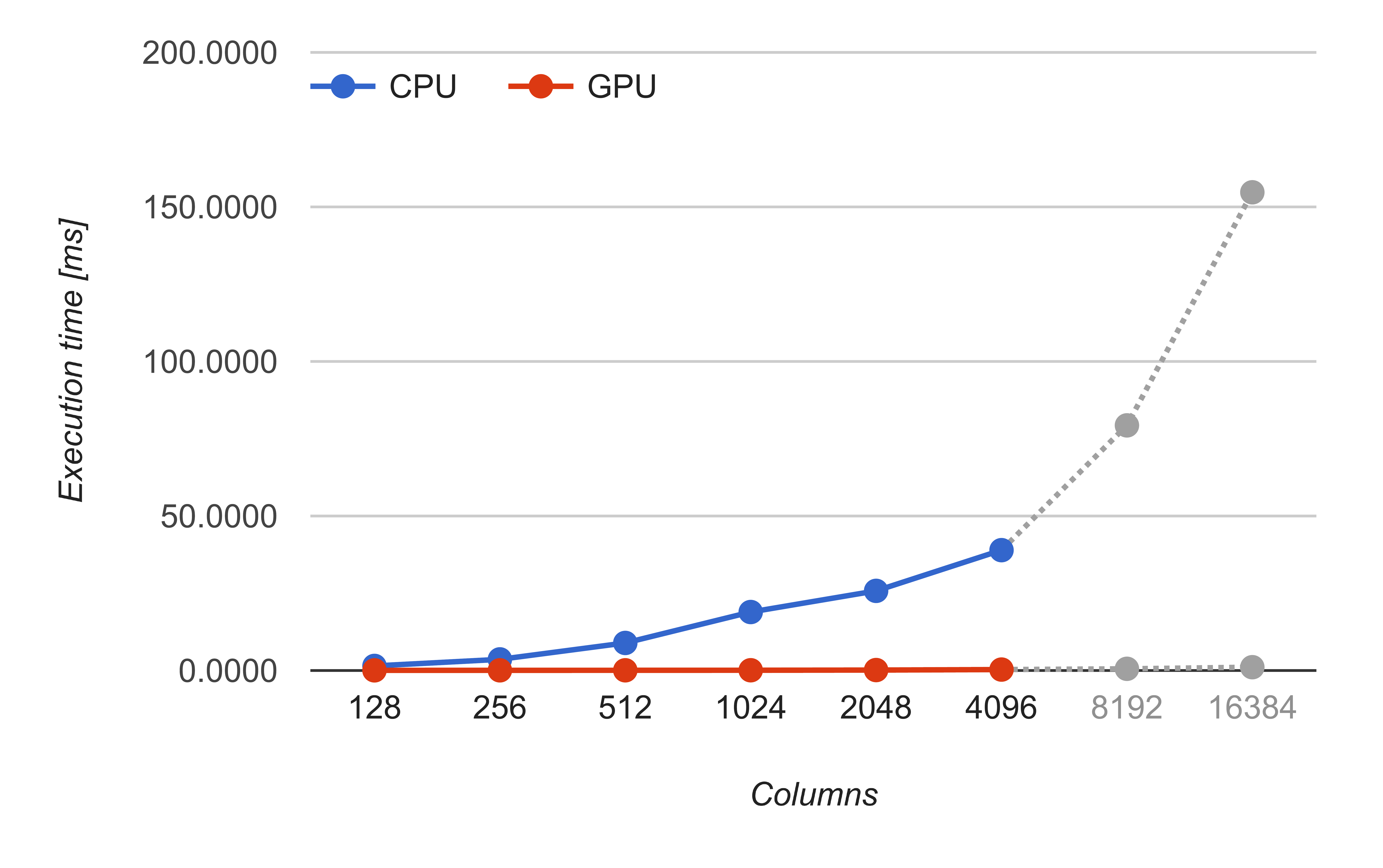}
    \caption{Average kernel exec time (with forecast)}
\end{subfigure}
\begin{subfigure}[t]{0.48\textwidth}
    \includegraphics[width=\textwidth]{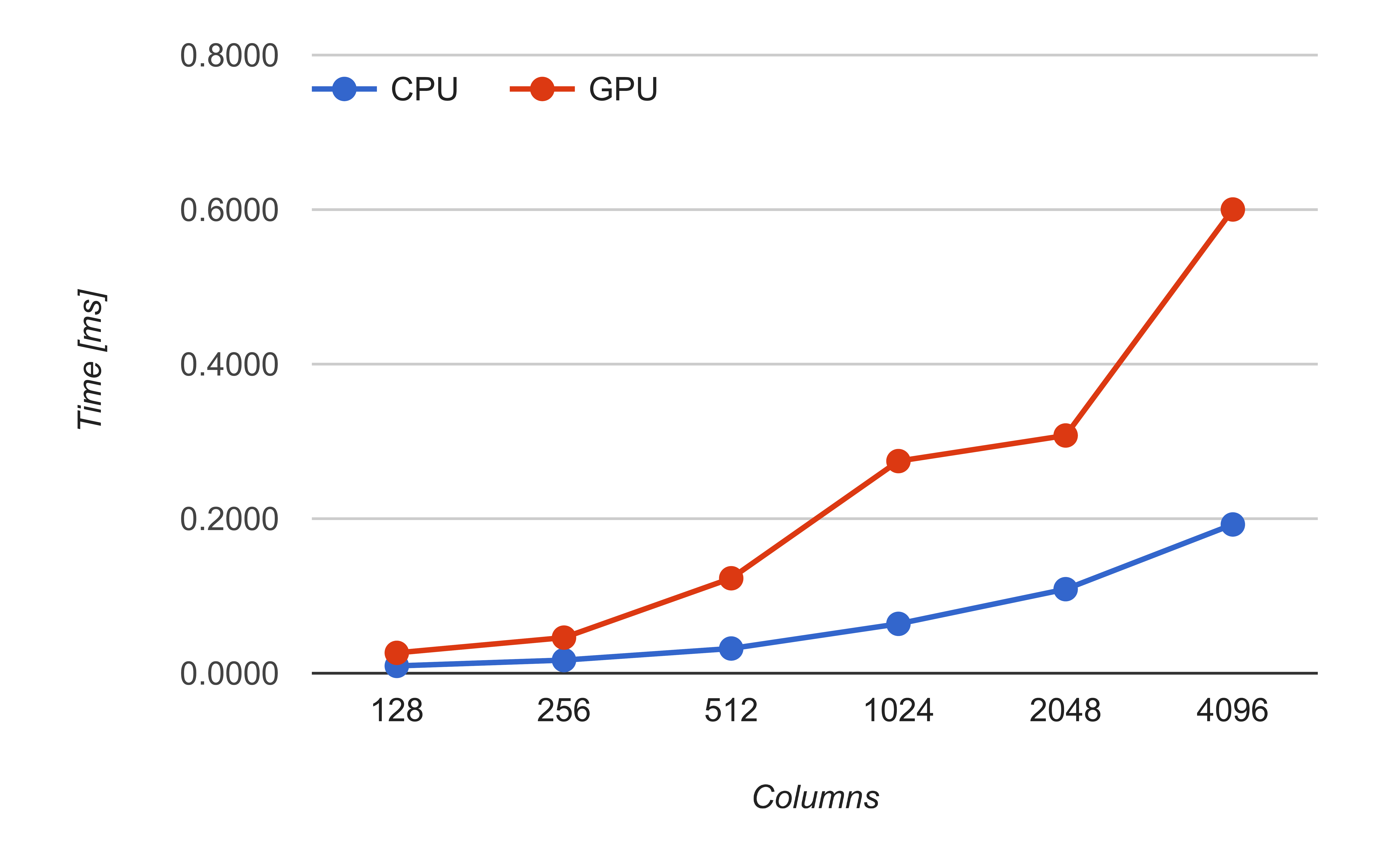}
    \caption{Average host--to--device data transfer time}
\end{subfigure}
\begin{subfigure}[t]{0.48\textwidth}
    \includegraphics[width=\textwidth]{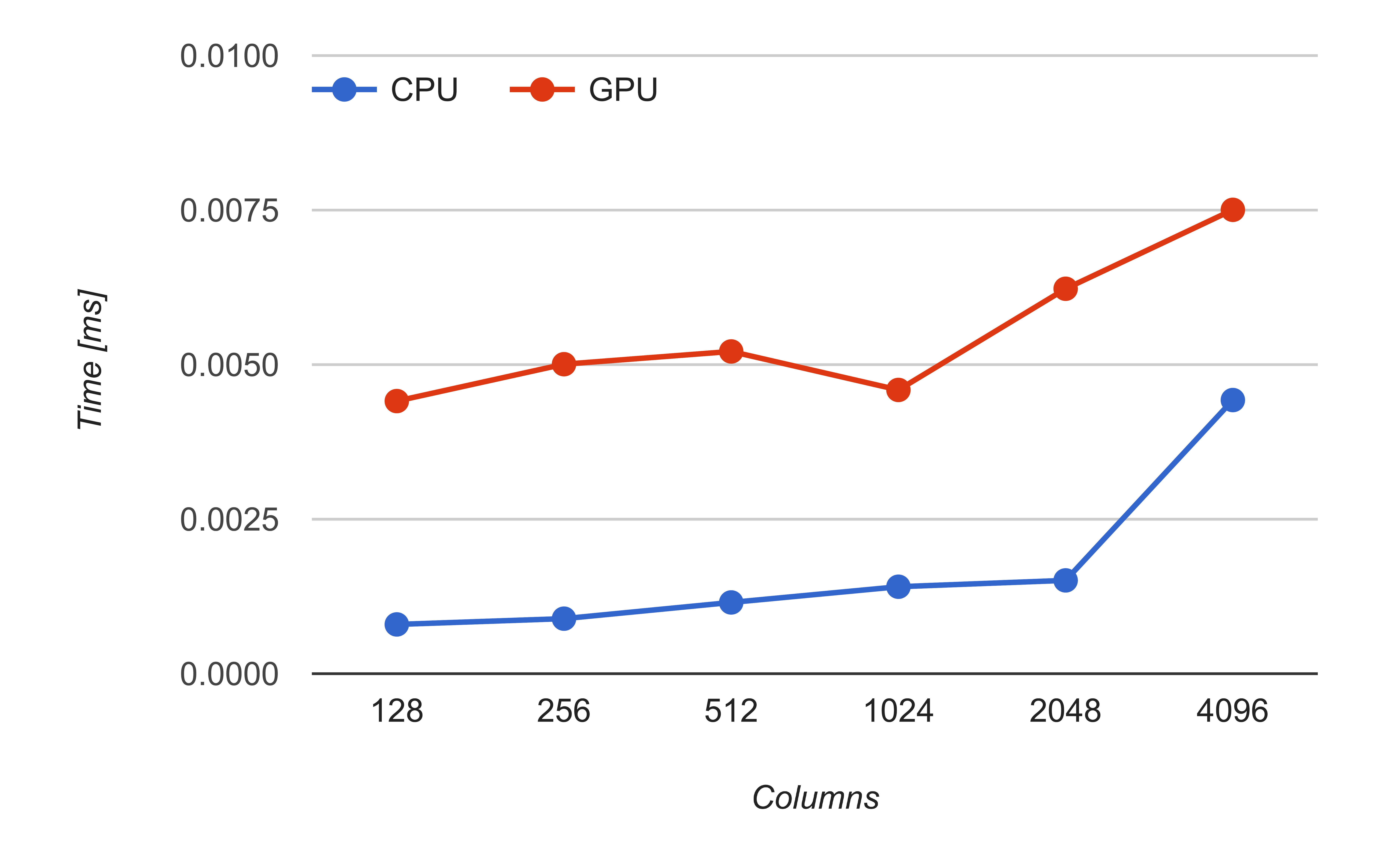}
    \caption{Average device--to--host data transfer time}
\end{subfigure}
\caption{Profiling results for columns}
\label{fig:pr_columns_time}
\end{figure}

\begin{figure}[p]
\centering
\begin{subfigure}{0.48\textwidth}
    \includegraphics[width=\textwidth]{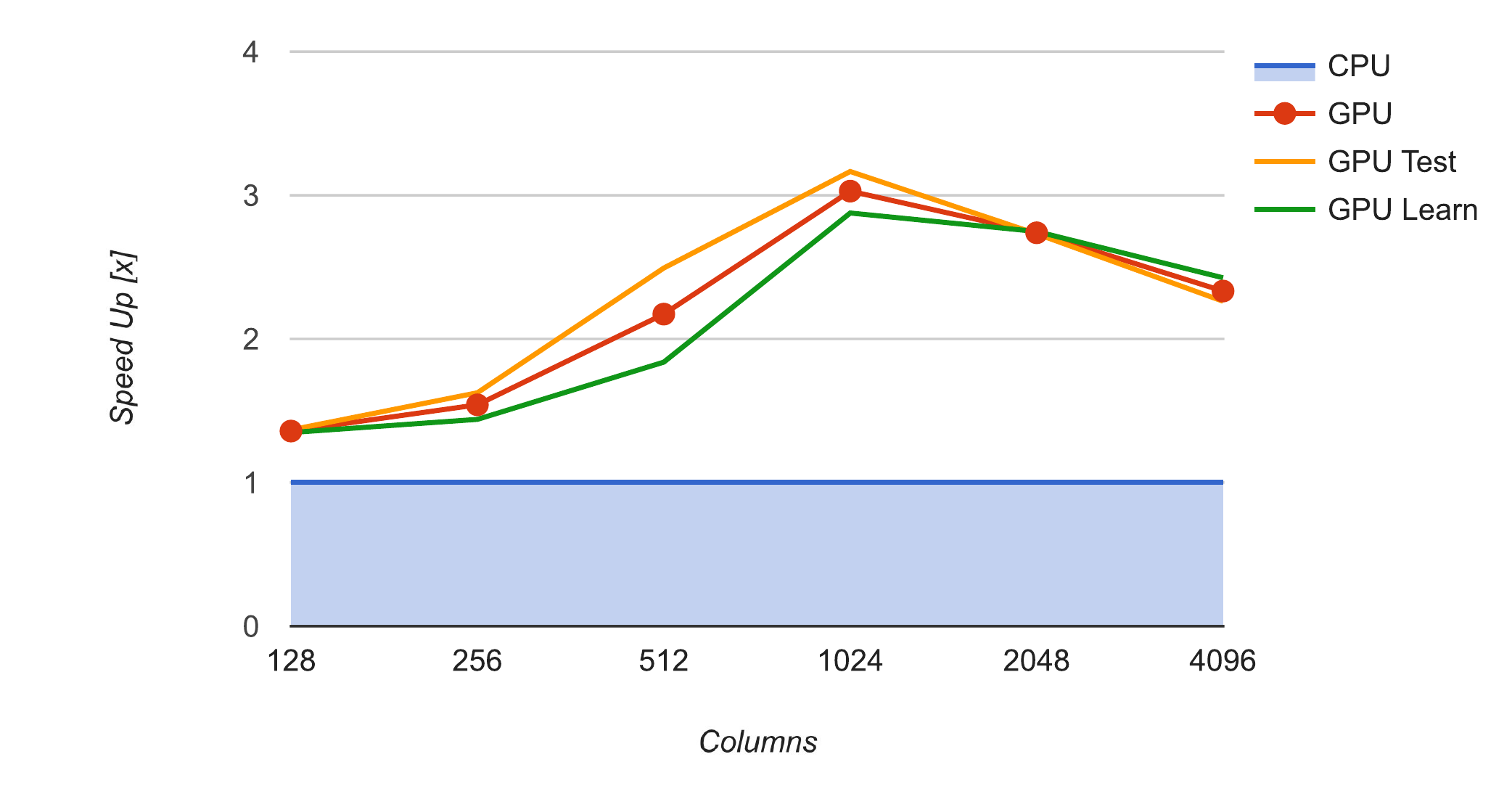}
    \caption{GPU vs CPU OCL}
\end{subfigure}
\begin{subfigure}{0.48\textwidth}
    \includegraphics[width=\textwidth]{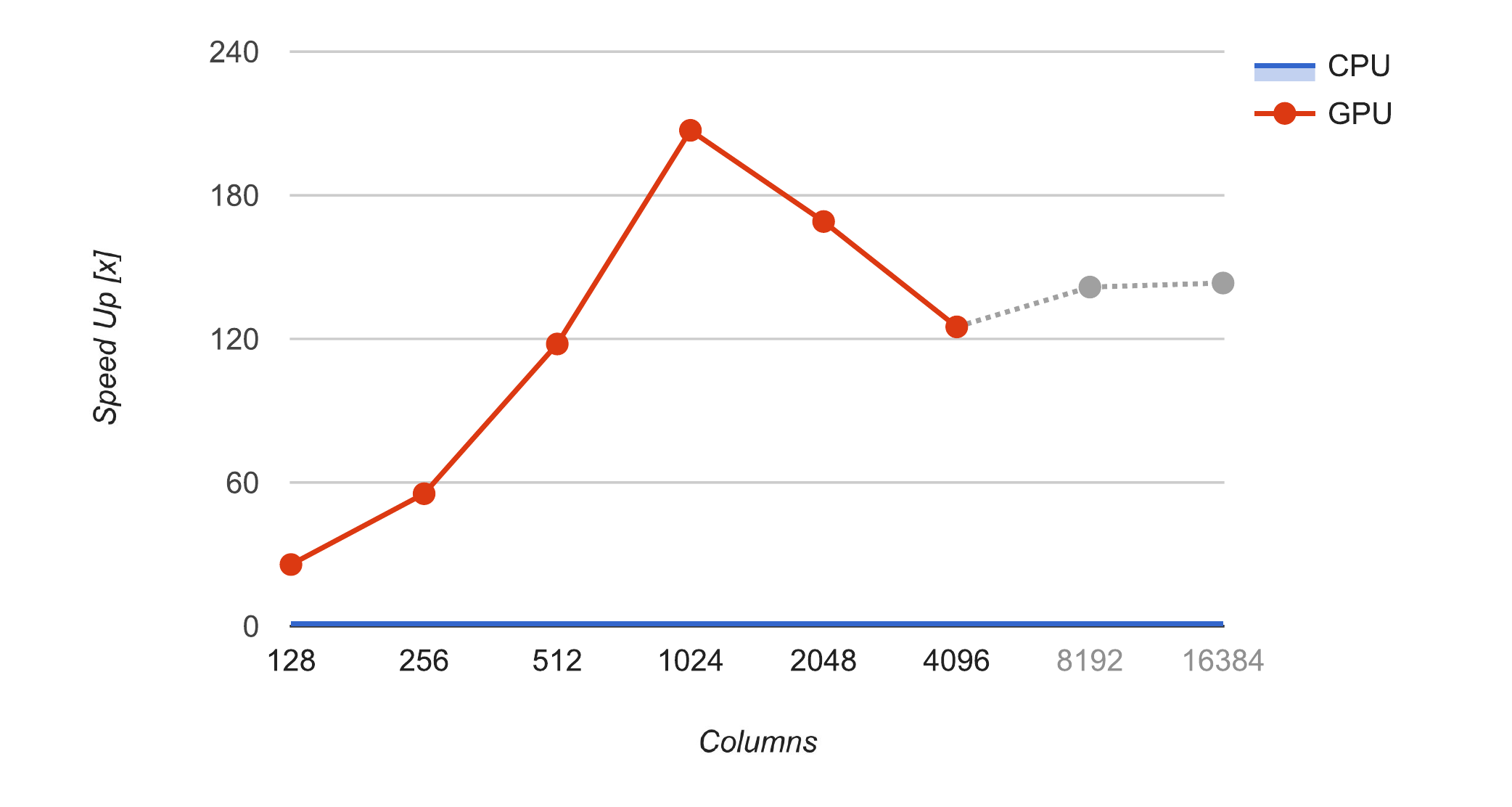}
    \caption{GPU vs CPU kernel (with forecast)}
\end{subfigure}
\begin{subfigure}{0.48\textwidth}
    \includegraphics[width=\textwidth]{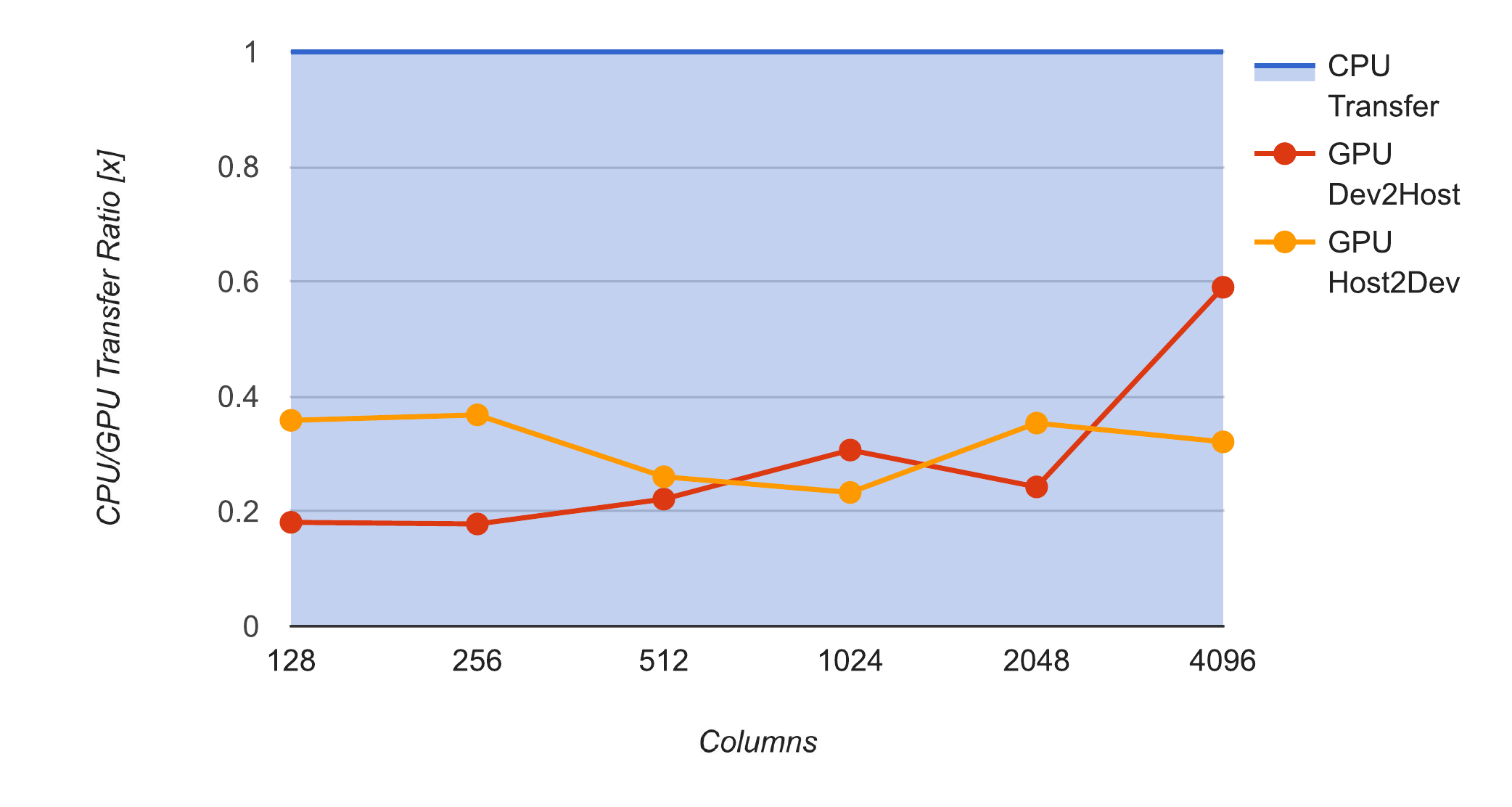}
    \caption{GPU vs CPU data transfer}
\end{subfigure}
\caption{Profiling results for columns}
\label{fig:pr_columns_speedup}
\end{figure}

\begin{figure}[p]
\centering
\begin{subfigure}{0.48\textwidth}
    \includegraphics[width=\textwidth]{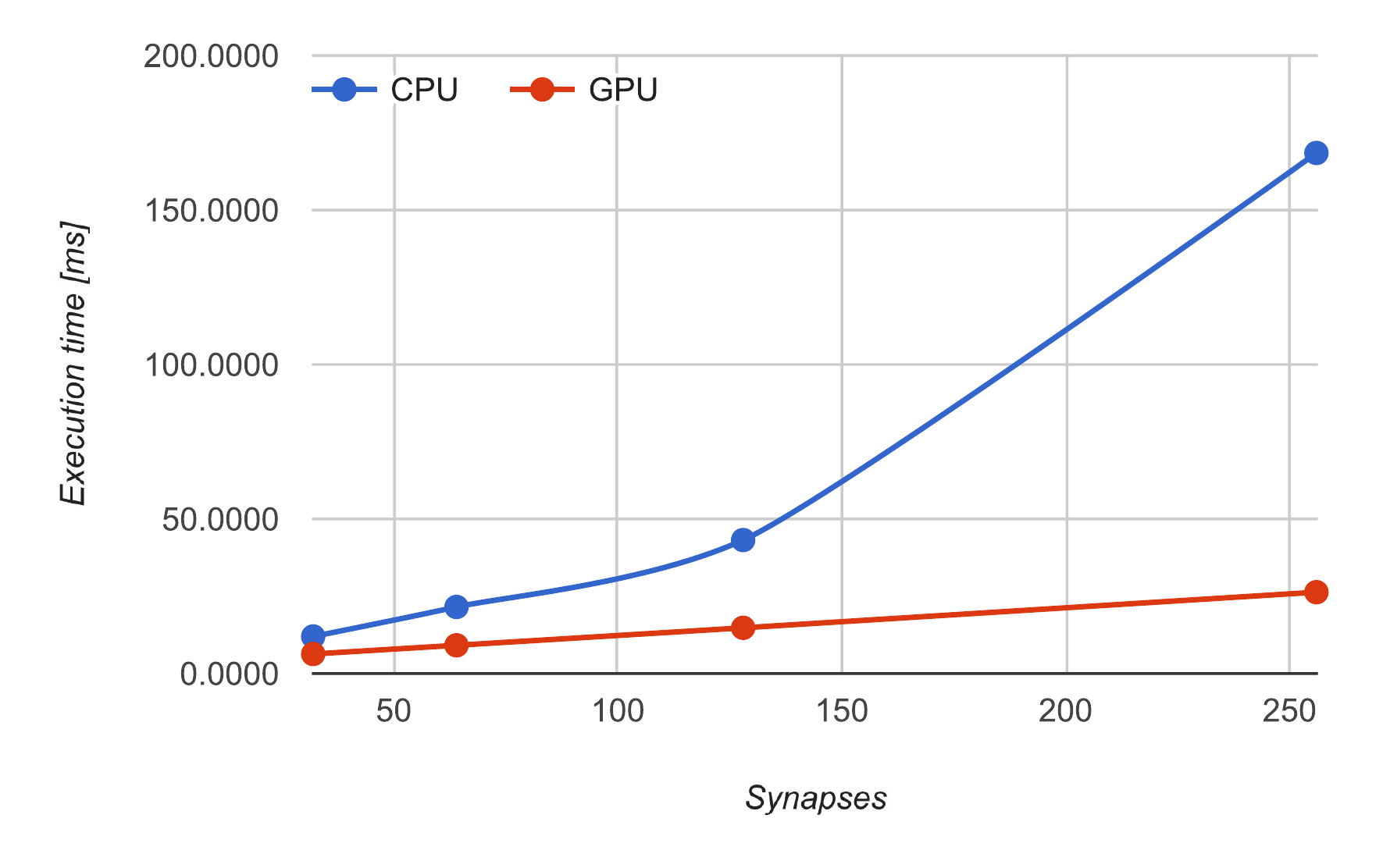}
    \caption{Average OCL kernel exec time}
\end{subfigure}
\begin{subfigure}{0.48\textwidth}
    \includegraphics[width=\textwidth]{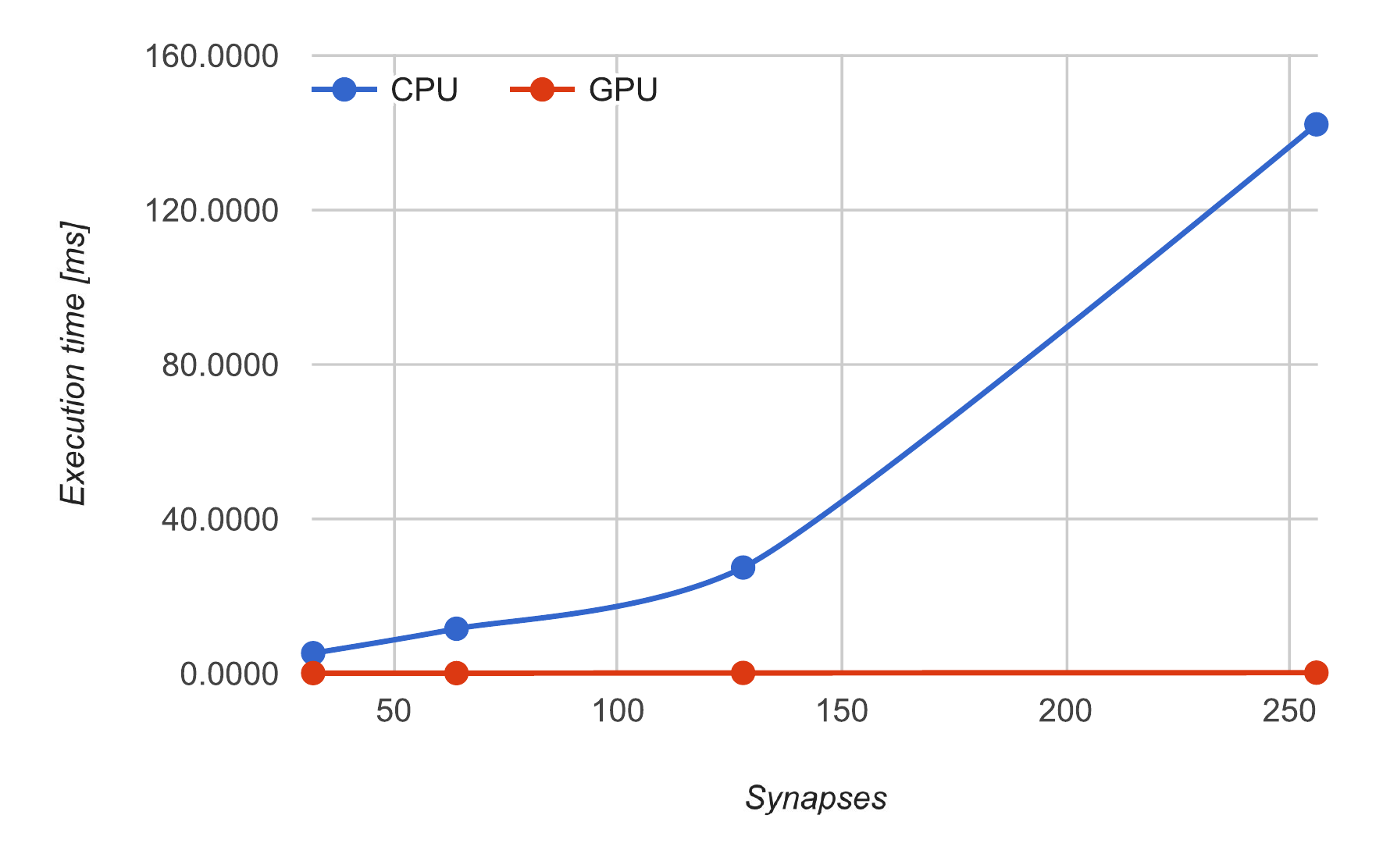}
    \caption{Average kernel exec time}
\end{subfigure}
\begin{subfigure}{0.48\textwidth}
    \includegraphics[width=\textwidth]{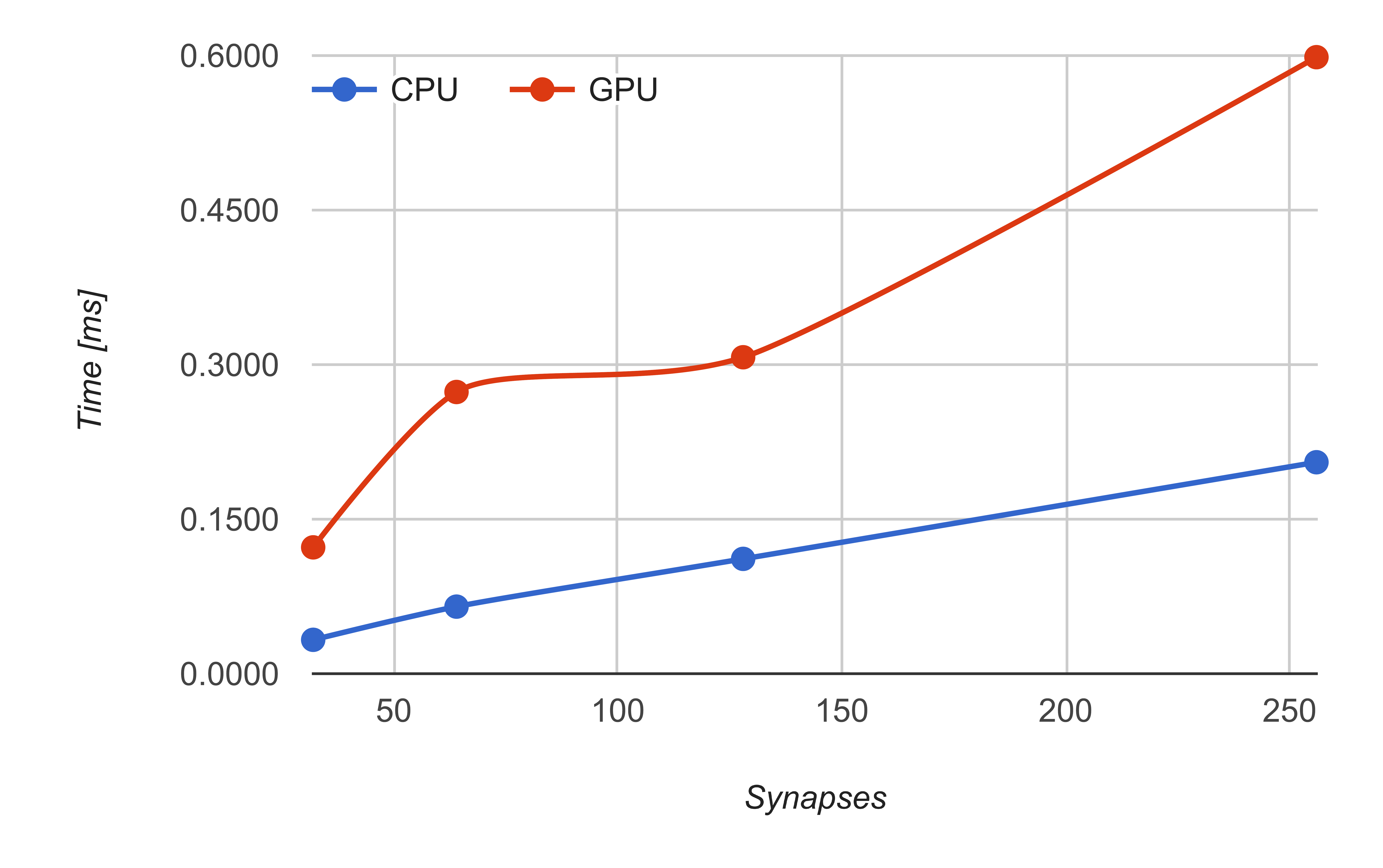}
    \caption{Average host--to--device data transfer time}
\end{subfigure}
\begin{subfigure}{0.48\textwidth}
    \includegraphics[width=\textwidth]{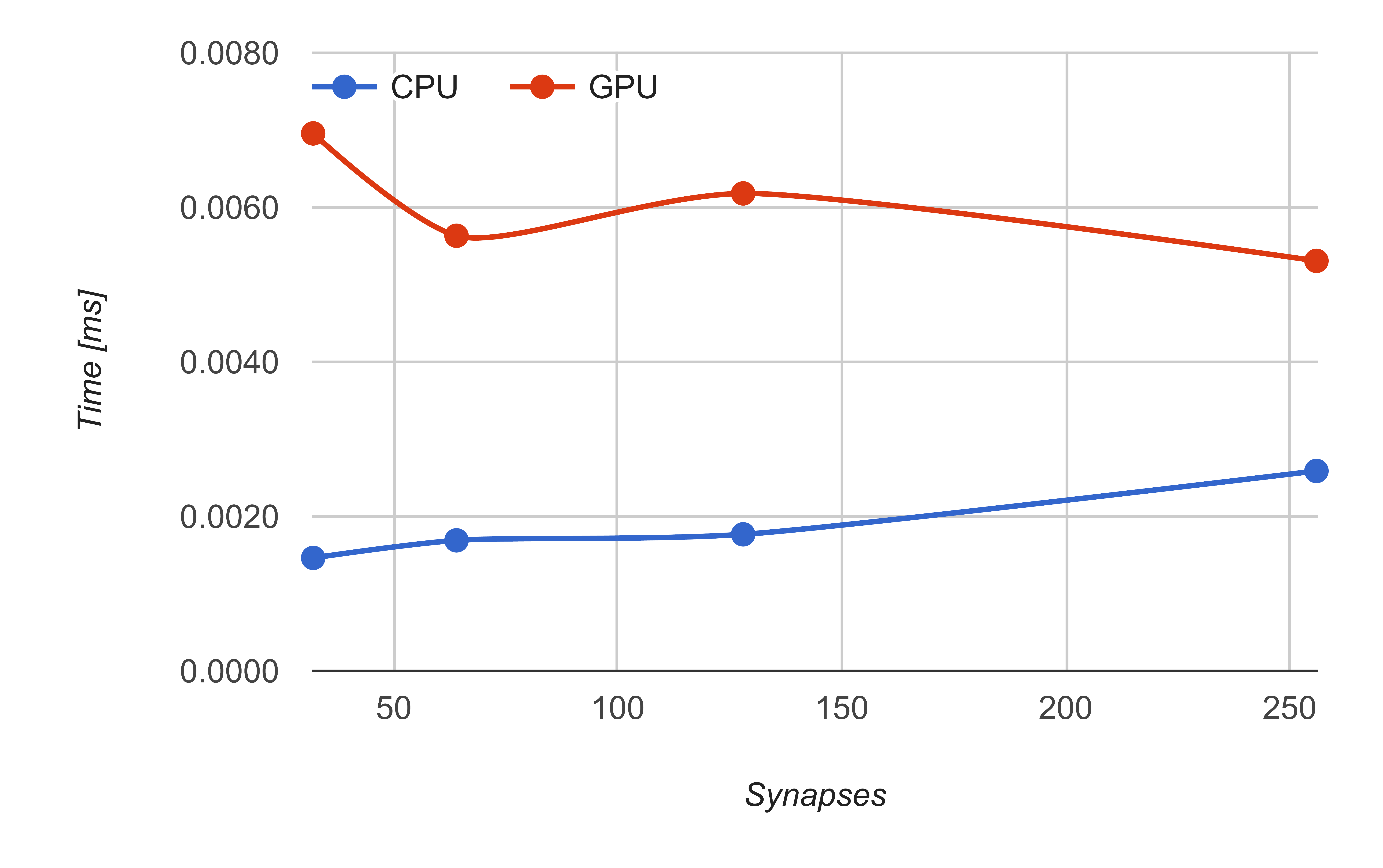}
    \caption{Average device--to--host data transfer time}
\end{subfigure}
\caption{Profiling results for synapses}
\label{fig:pr_synapses_time}
\end{figure}

\begin{figure}[p]
\centering
\begin{subfigure}{0.48\textwidth}
    \includegraphics[width=\textwidth]{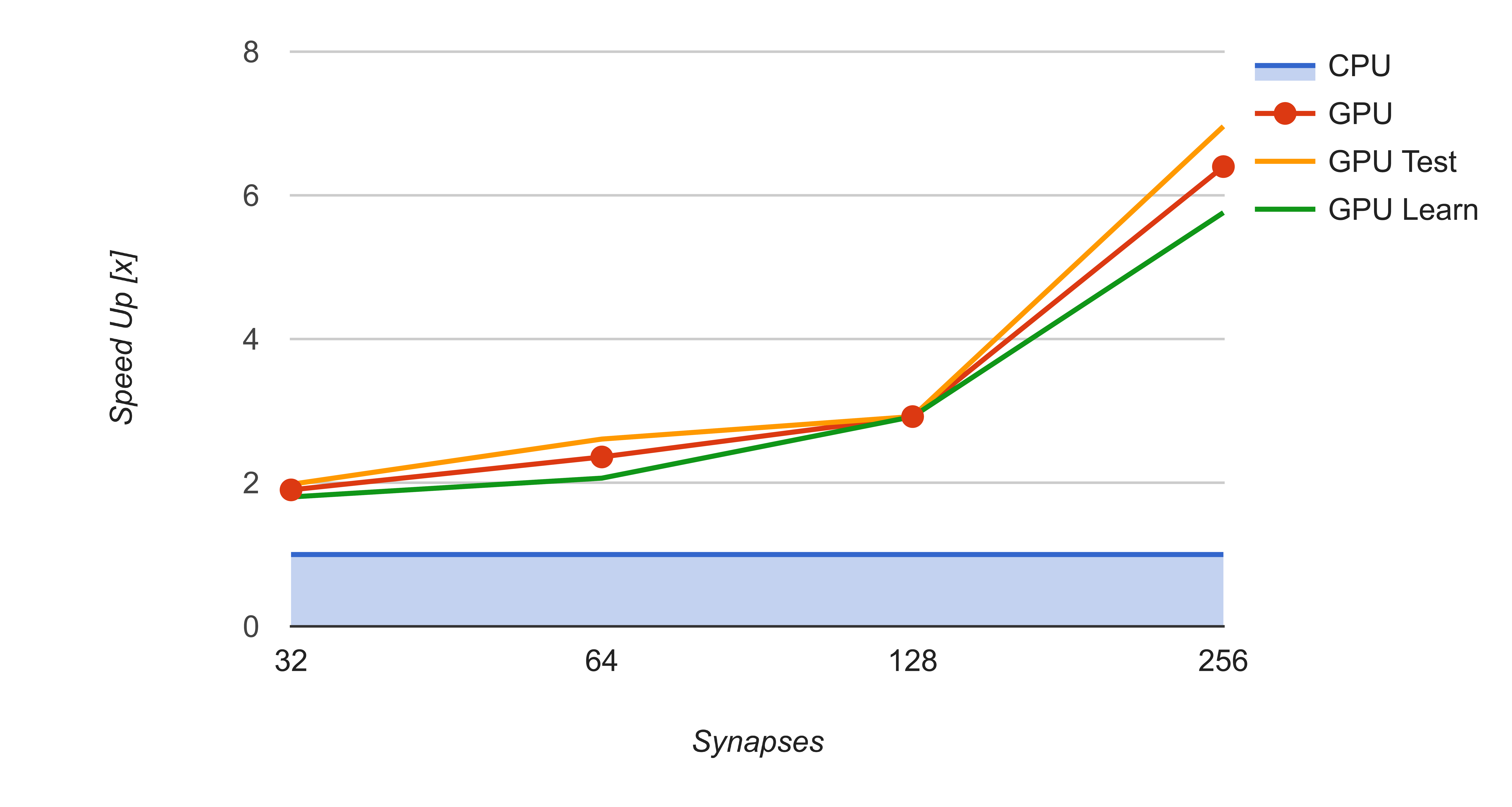}
    \caption{GPU vs CPU OCL}
\end{subfigure}
\begin{subfigure}{0.48\textwidth}
    \includegraphics[width=\textwidth]{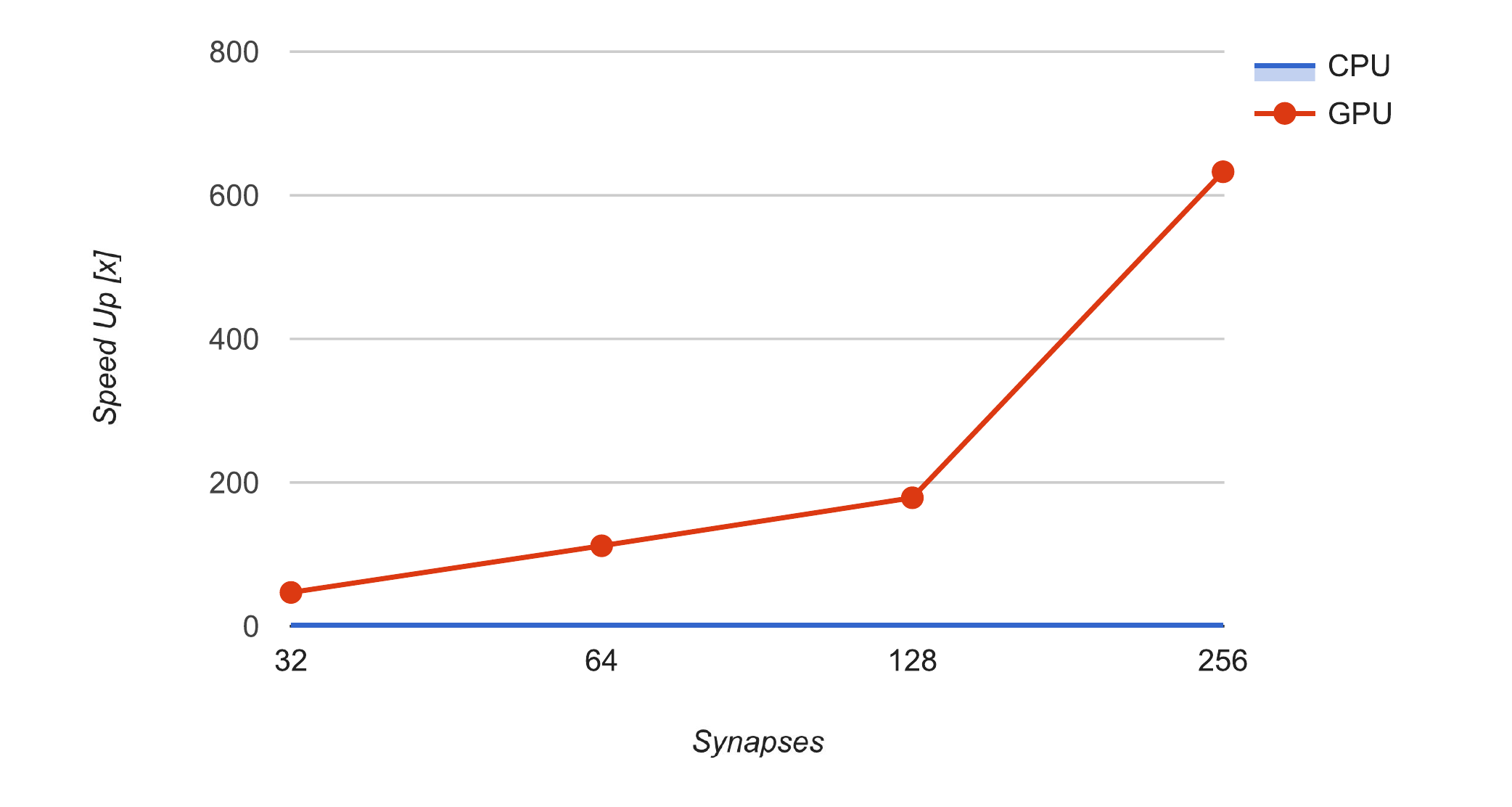}
    \caption{GPU vs CPU kernel}
\end{subfigure}
\begin{subfigure}{0.48\textwidth}
    \includegraphics[width=\textwidth]{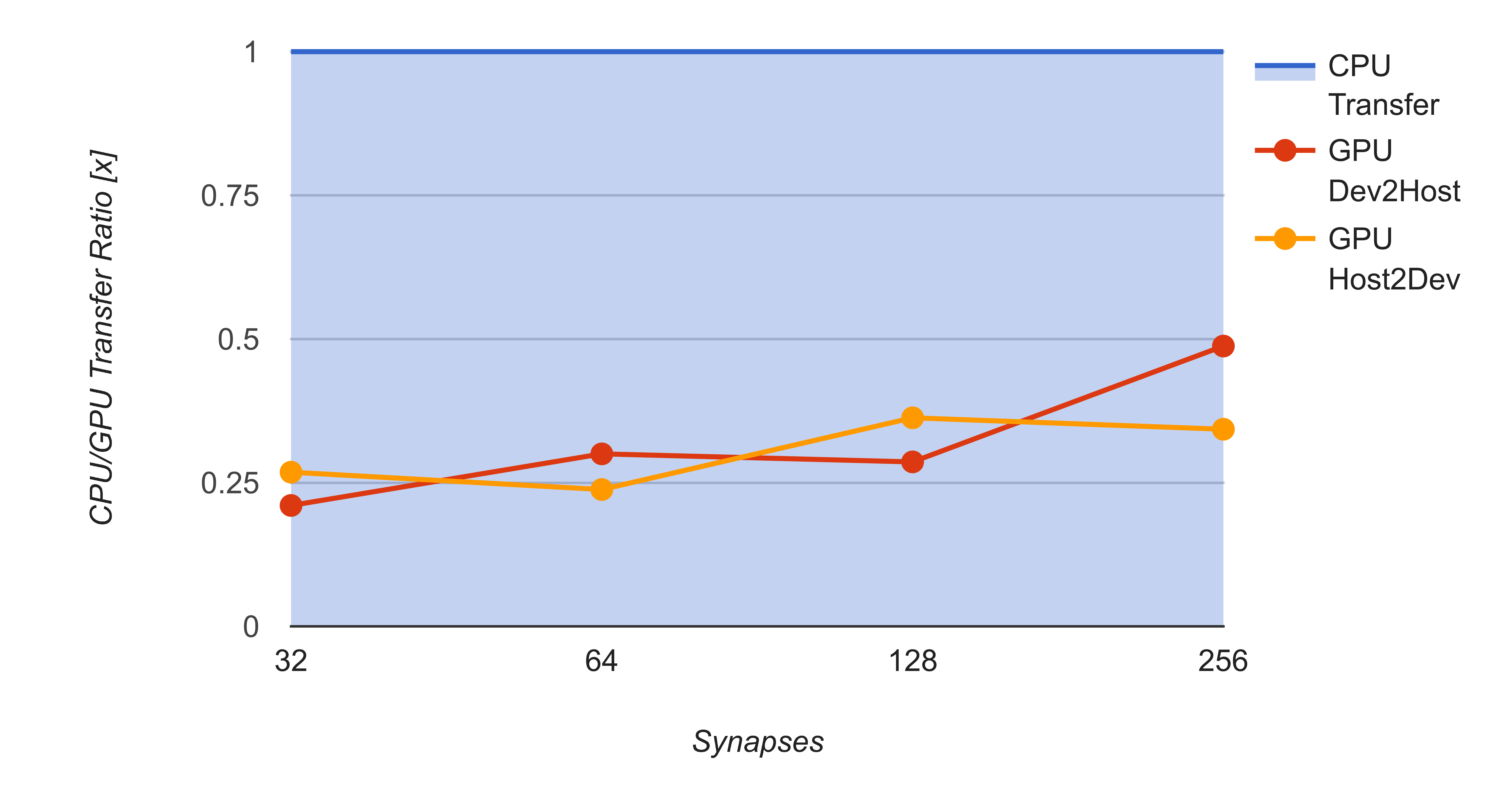}
    \caption{GPU vs CPU data transfer}
\end{subfigure}
\caption{Profiling results for synapses}
\label{fig:pr_synapses_speedup}
\end{figure}

\begin{figure}[p]
\centering
\begin{subfigure}{0.48\textwidth}
    \includegraphics[width=\textwidth]{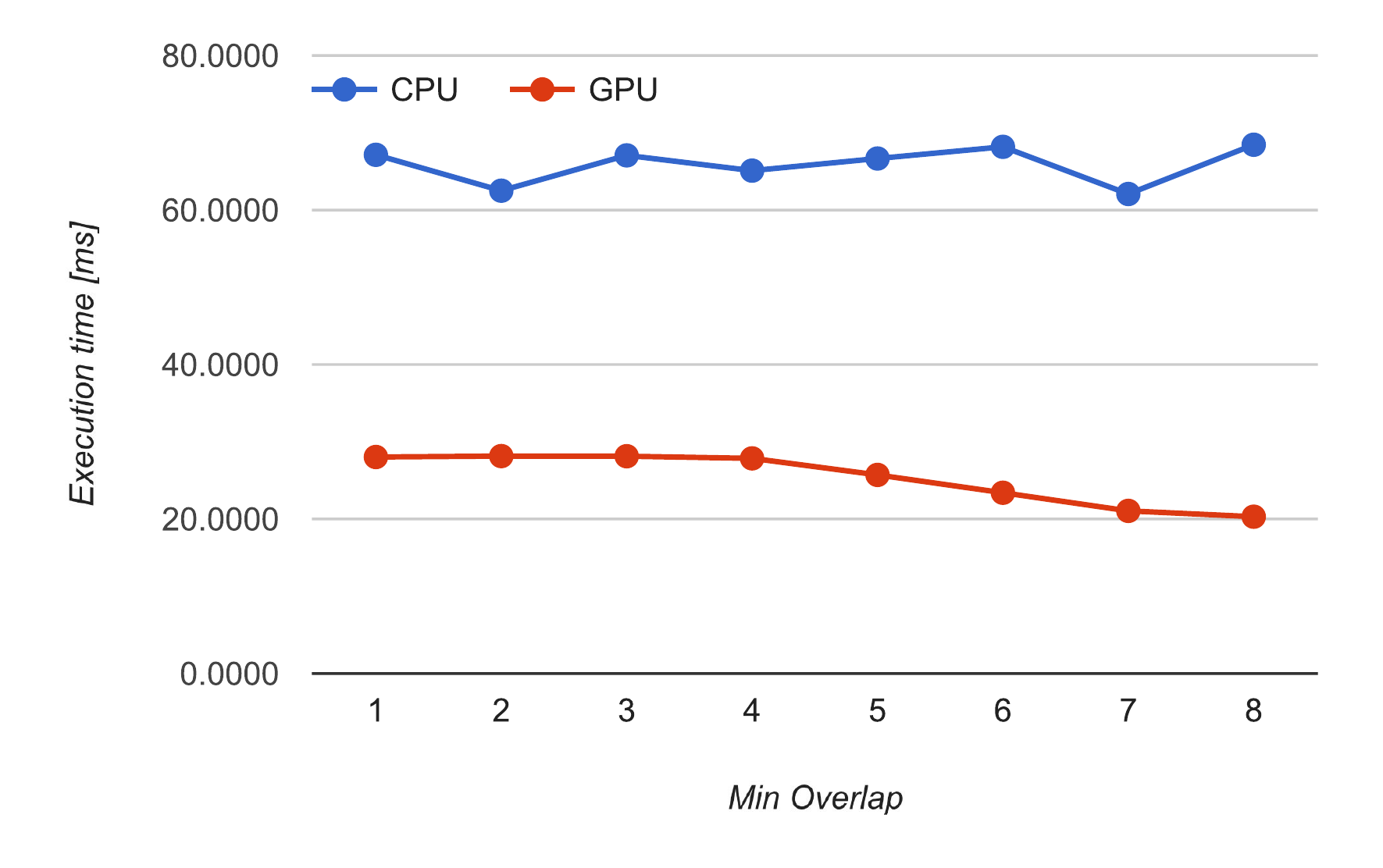}
    \caption{Average OCL kernel exec time}
\end{subfigure}
\begin{subfigure}{0.48\textwidth}
    \includegraphics[width=\textwidth]{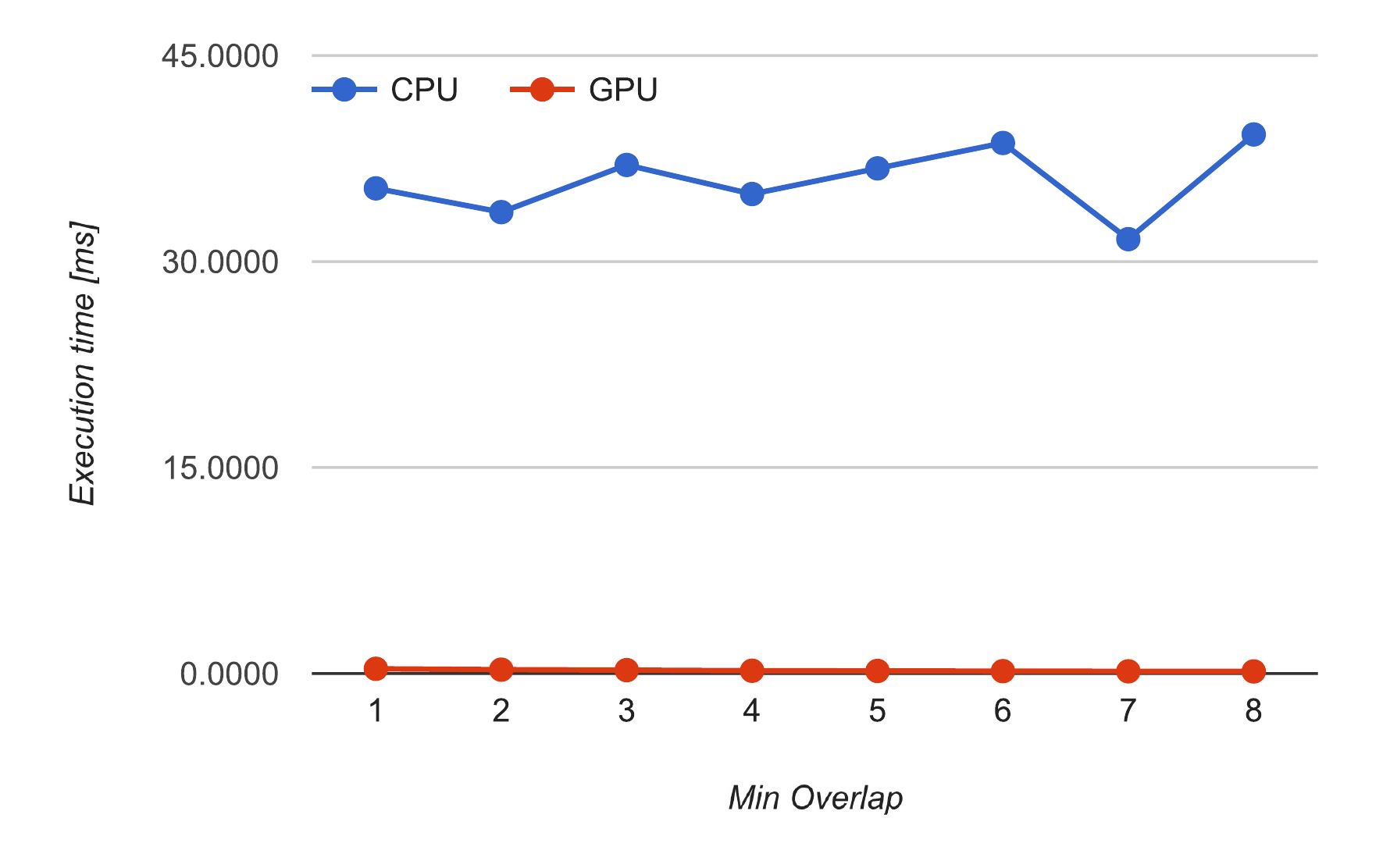}
    \caption{Average kernel exec time}
\end{subfigure}
\begin{subfigure}{0.48\textwidth}
    \includegraphics[width=\textwidth]{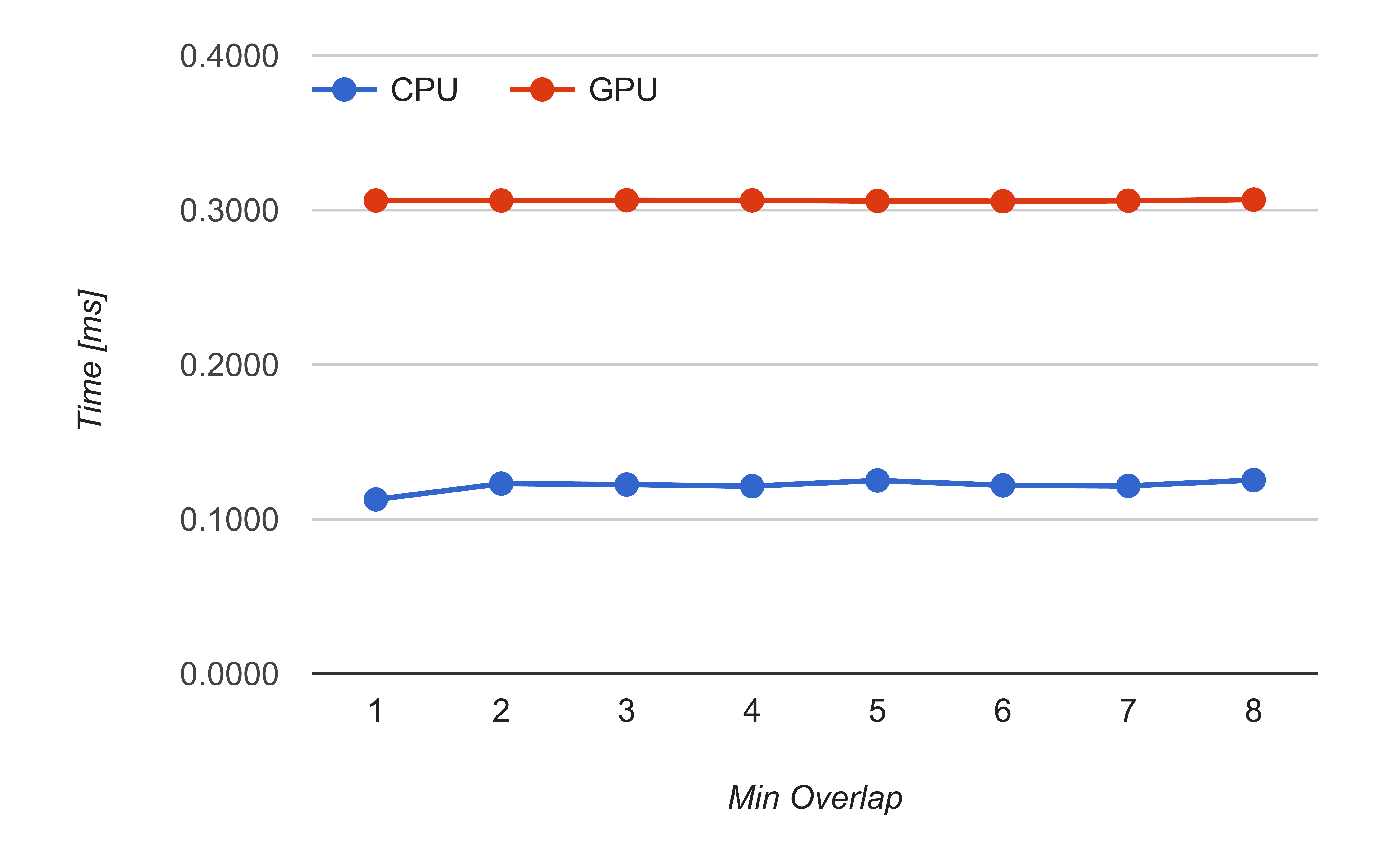}
    \caption{Average host--to--device data transfer time}
\end{subfigure}
\begin{subfigure}{0.48\textwidth}
    \includegraphics[width=\textwidth]{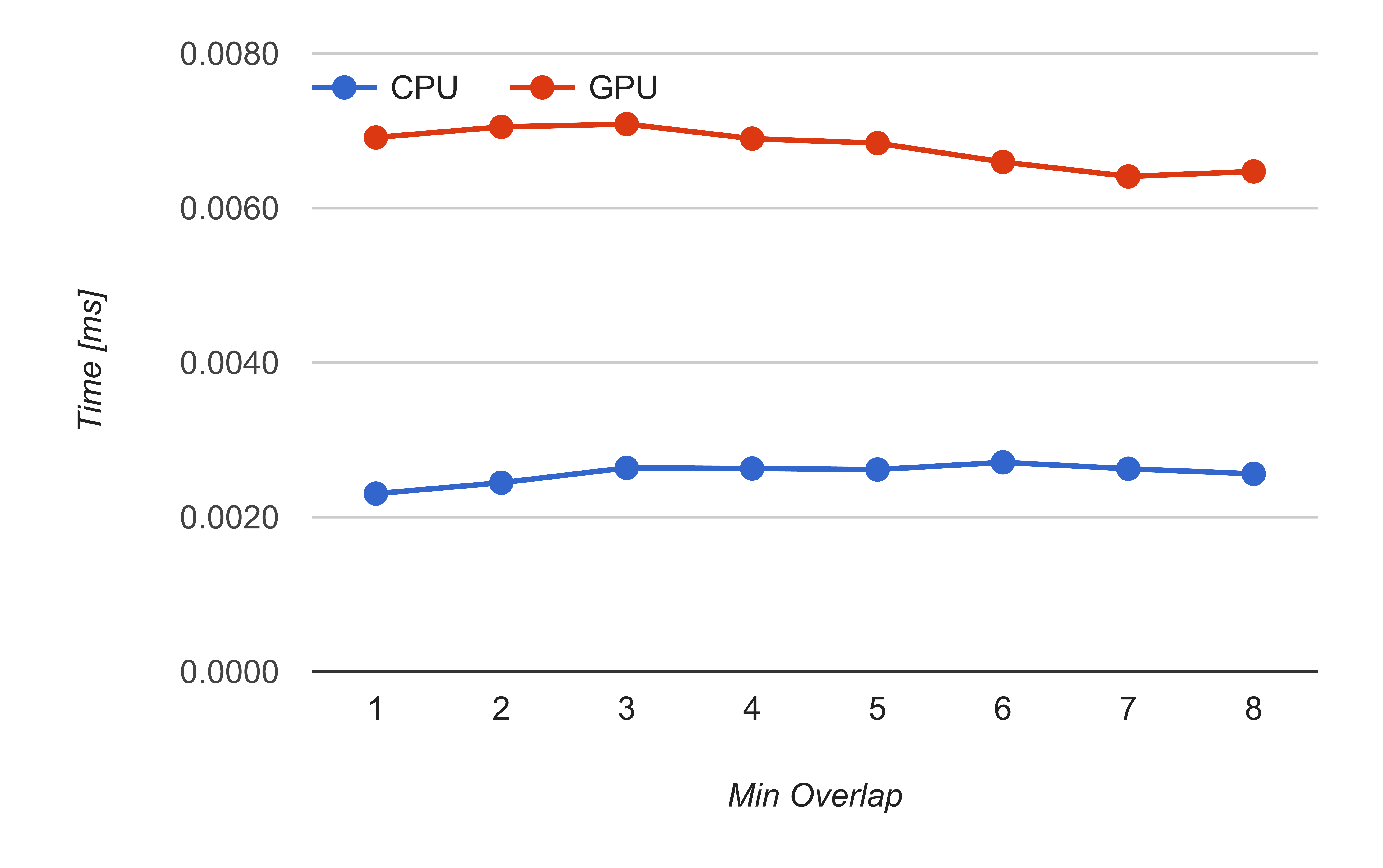}
    \caption{Average device--to--host data transfer time}
\end{subfigure}
\caption{Profiling results for min overlap}
\label{fig:pr_min_overlap_time}
\end{figure}

\begin{figure}[p]
\centering
\begin{subfigure}{0.48\textwidth}
    \includegraphics[width=\textwidth]{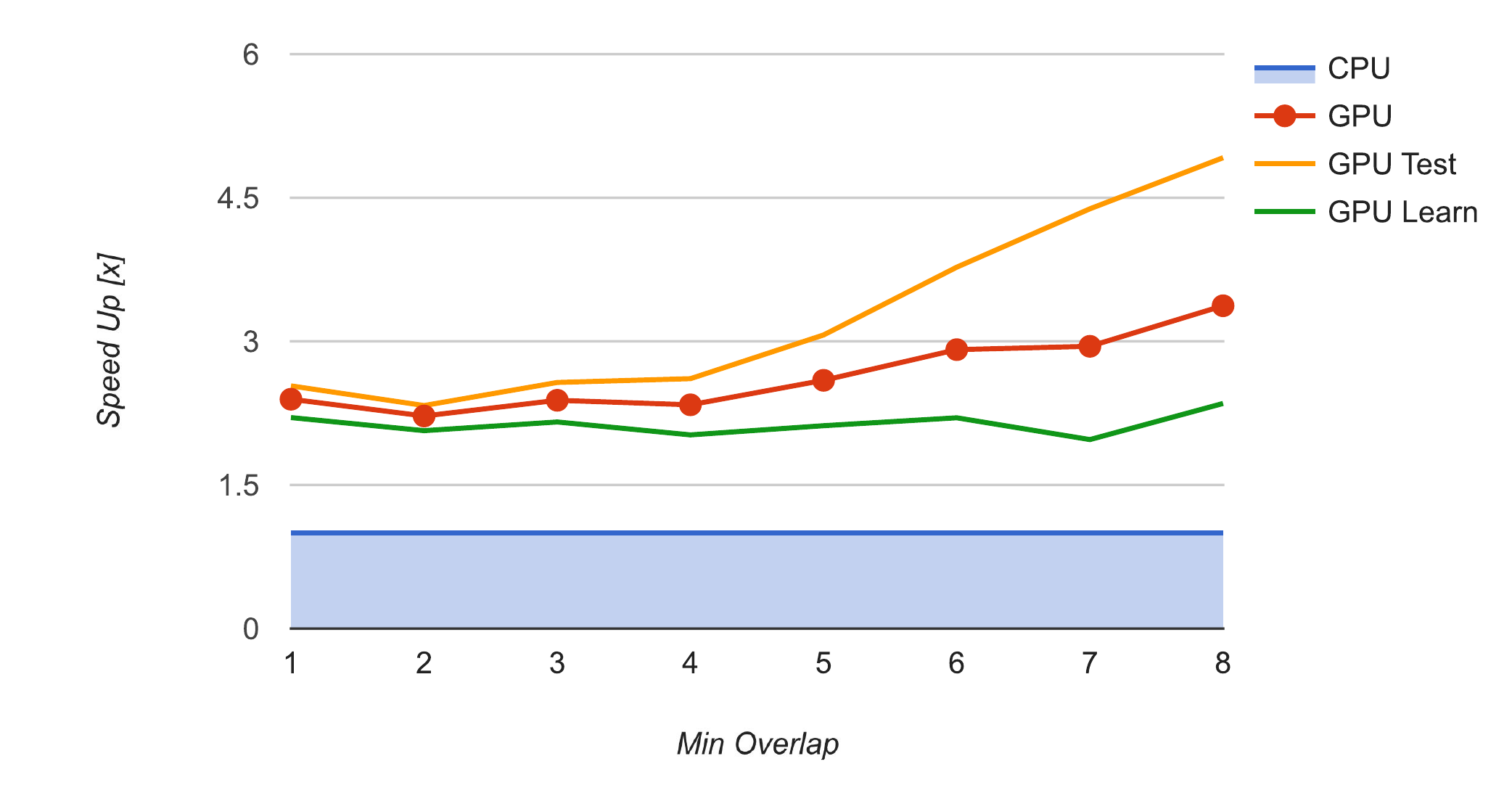}
    \caption{GPU vs CPU OCL}
\end{subfigure}
\begin{subfigure}{0.48\textwidth}
    \includegraphics[width=\textwidth]{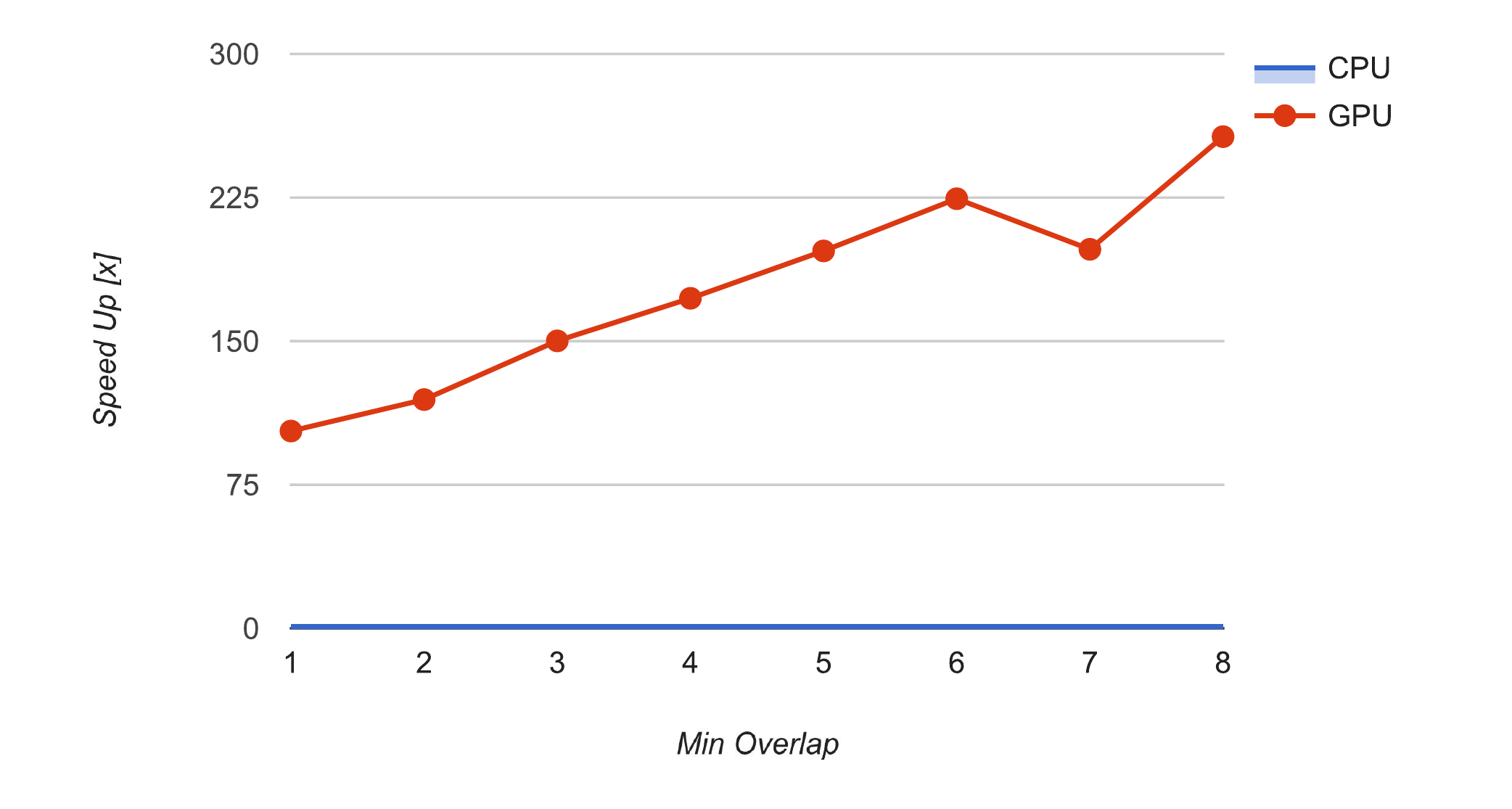}
    \caption{GPU vs CPU kernel}
\end{subfigure}
\begin{subfigure}{0.48\textwidth}
    \includegraphics[width=\textwidth]{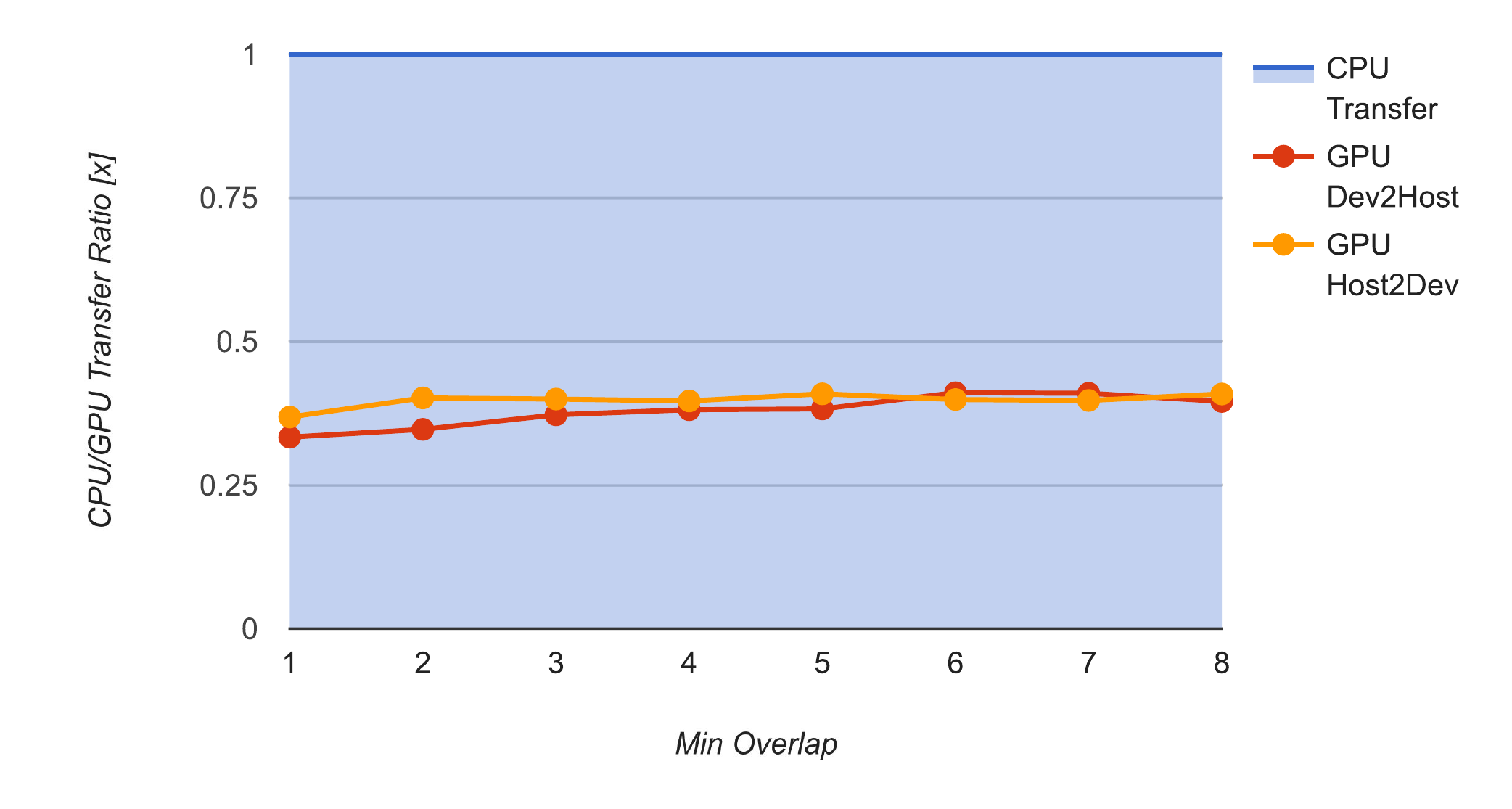}
    \caption{GPU vs CPU data transfer}
\end{subfigure}
\caption{Profiling results for min overlap}
\label{fig:pr_min_overlap_speedup}
\end{figure}

\begin{figure}[p]
\centering
\begin{subfigure}{0.48\textwidth}
    \includegraphics[width=\textwidth]{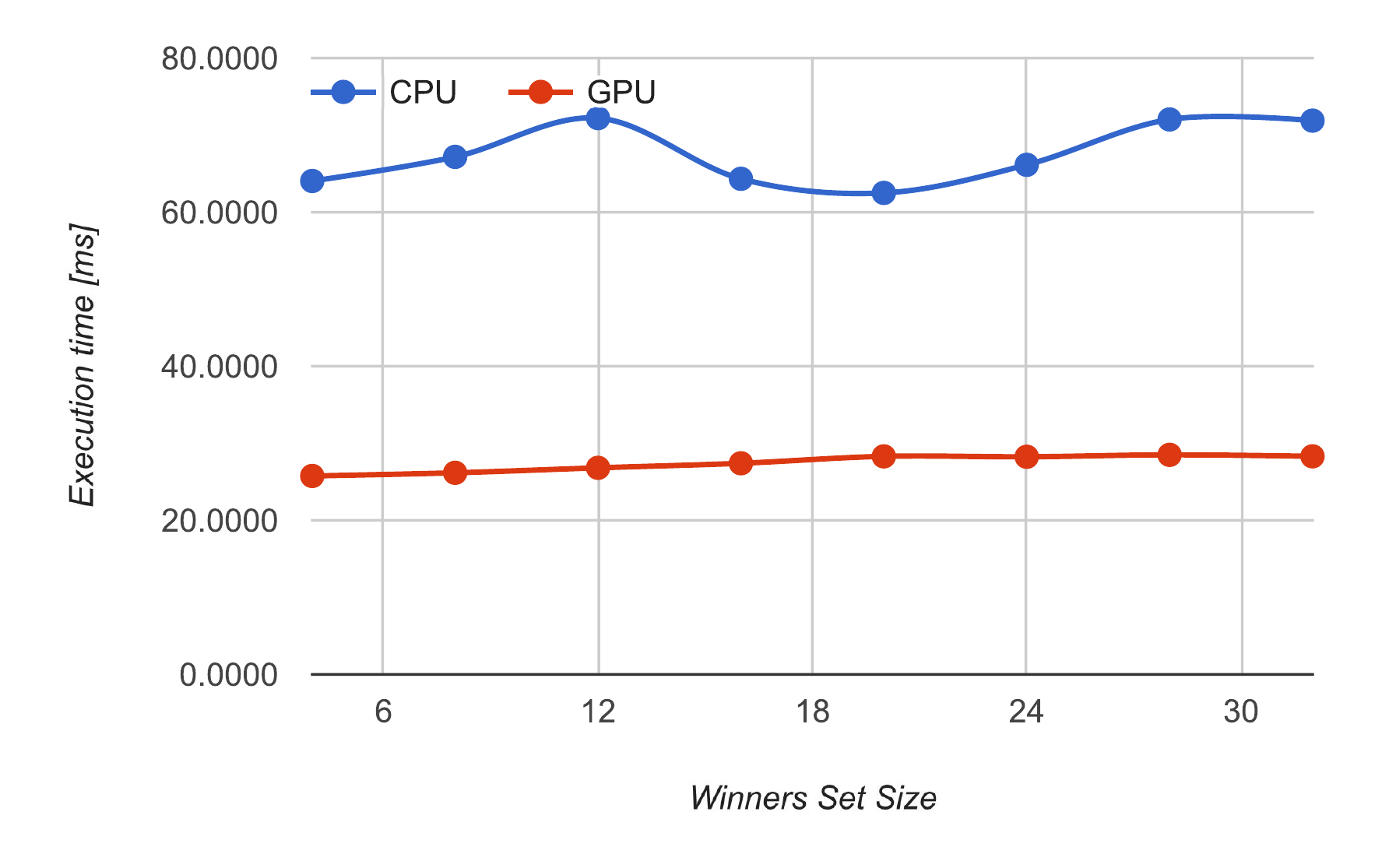}
    \caption{Average OCL kernel exec time}
\end{subfigure}
\begin{subfigure}{0.48\textwidth}
    \includegraphics[width=\textwidth]{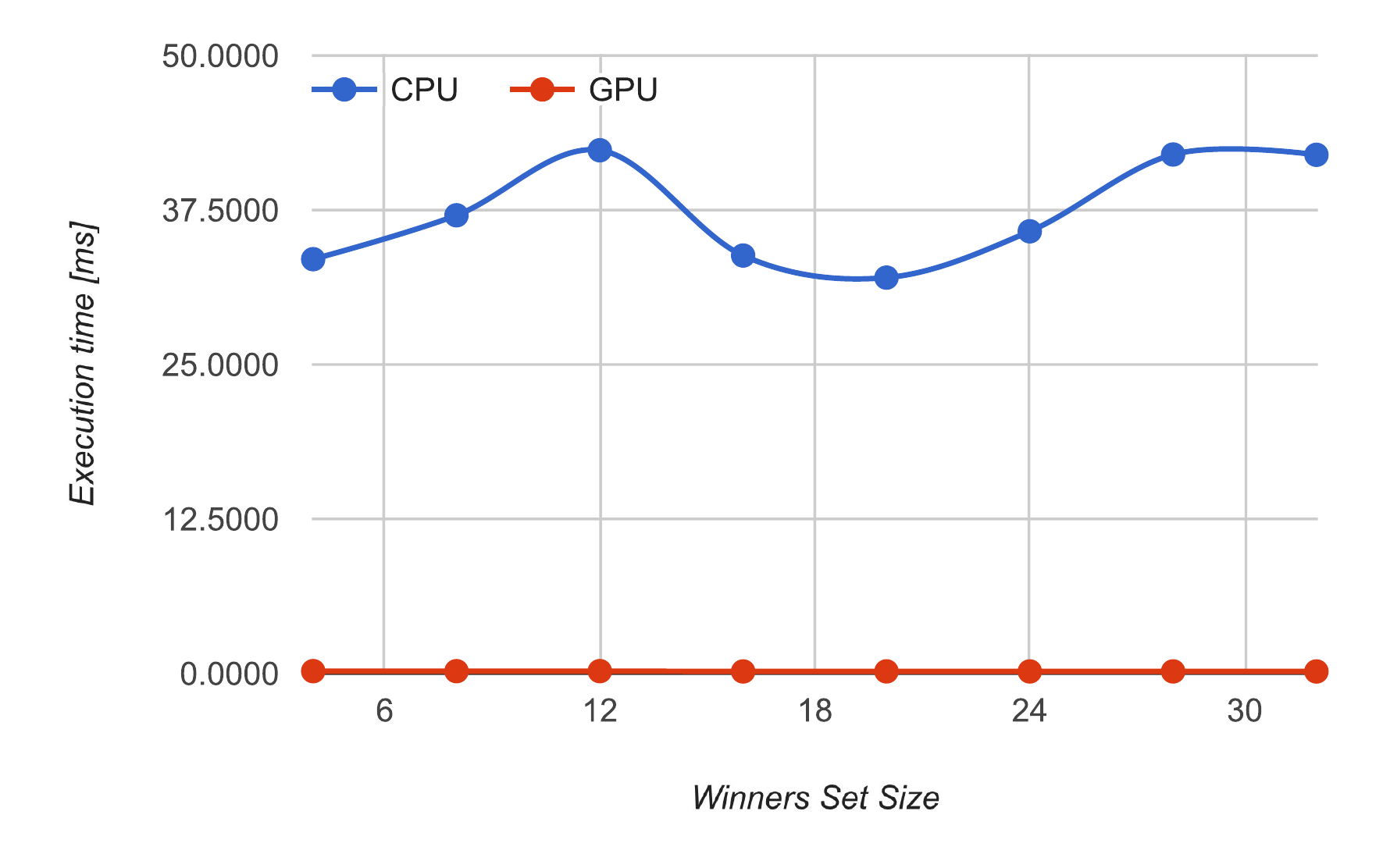}
    \caption{Average kernel exec time}
\end{subfigure}
\begin{subfigure}{0.48\textwidth}
    \includegraphics[width=\textwidth]{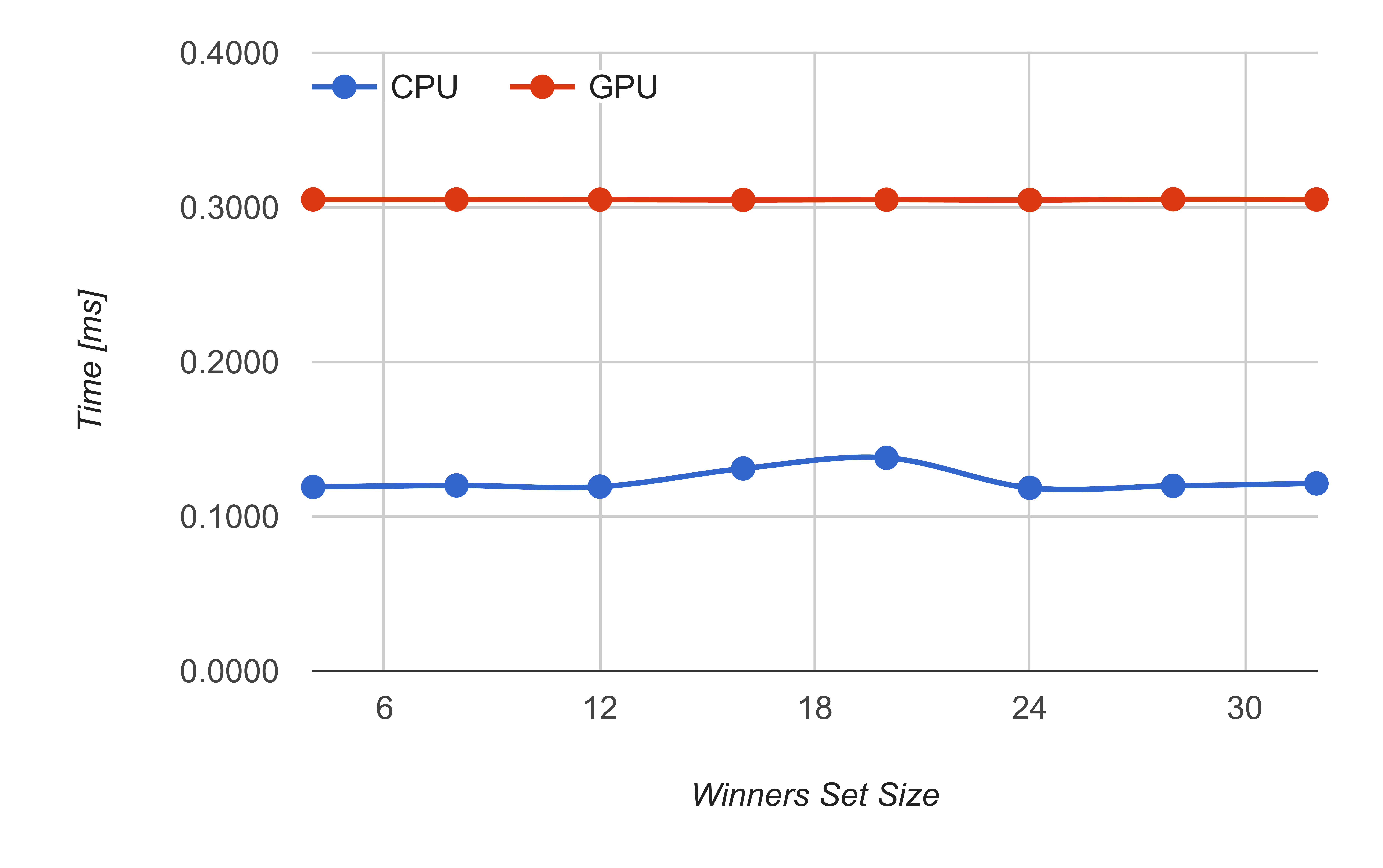}
    \caption{Average host--to--device data transfer time}
\end{subfigure}
\begin{subfigure}{0.48\textwidth}
    \includegraphics[width=\textwidth]{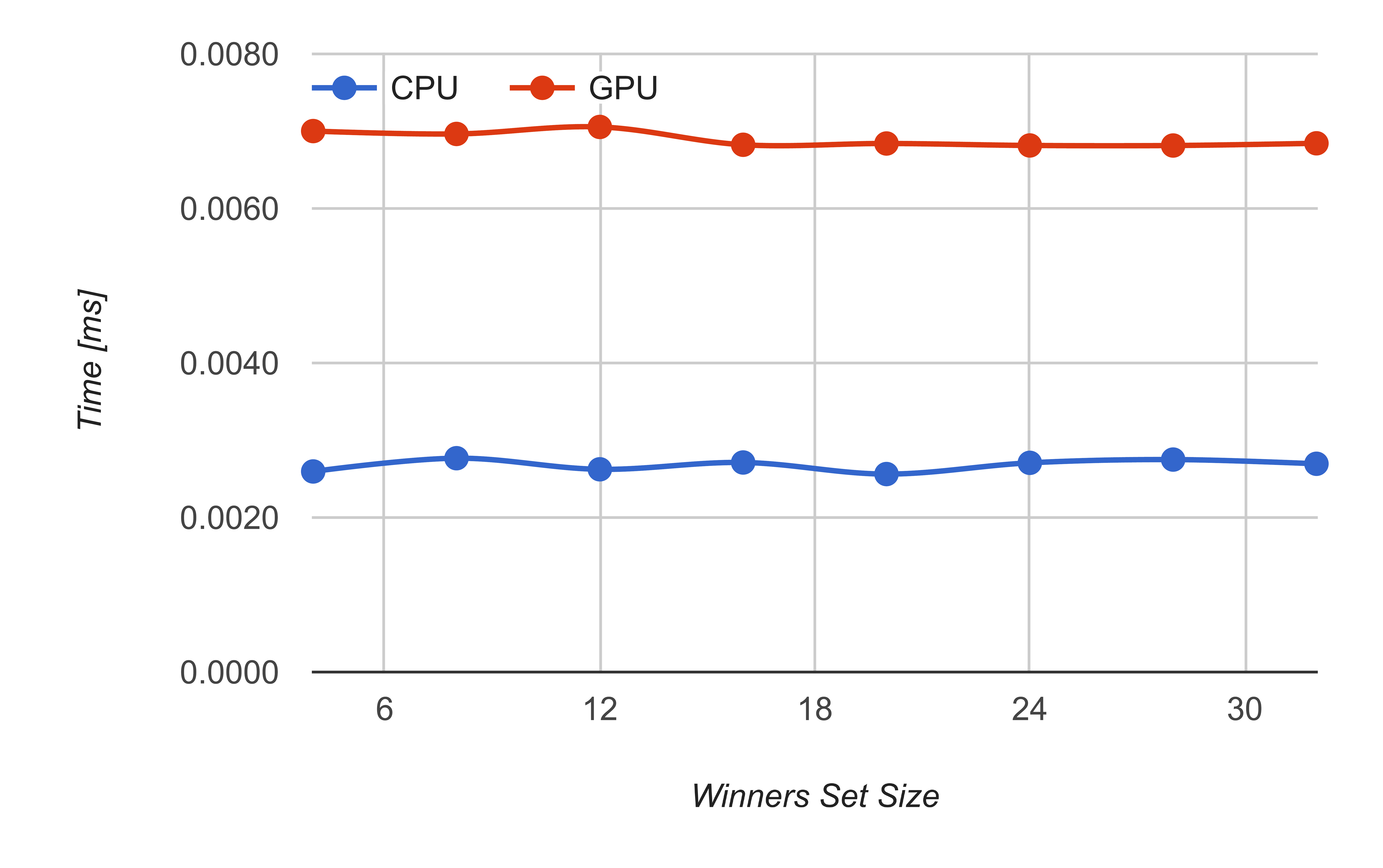}
    \caption{Average device--to--host data transfer time}
\end{subfigure}
\caption{Profiling results for winners set size}
\label{fig:pr_wss_time}
\end{figure}

\begin{figure}[p]
\centering
\begin{subfigure}{0.48\textwidth}
    \includegraphics[width=\textwidth]{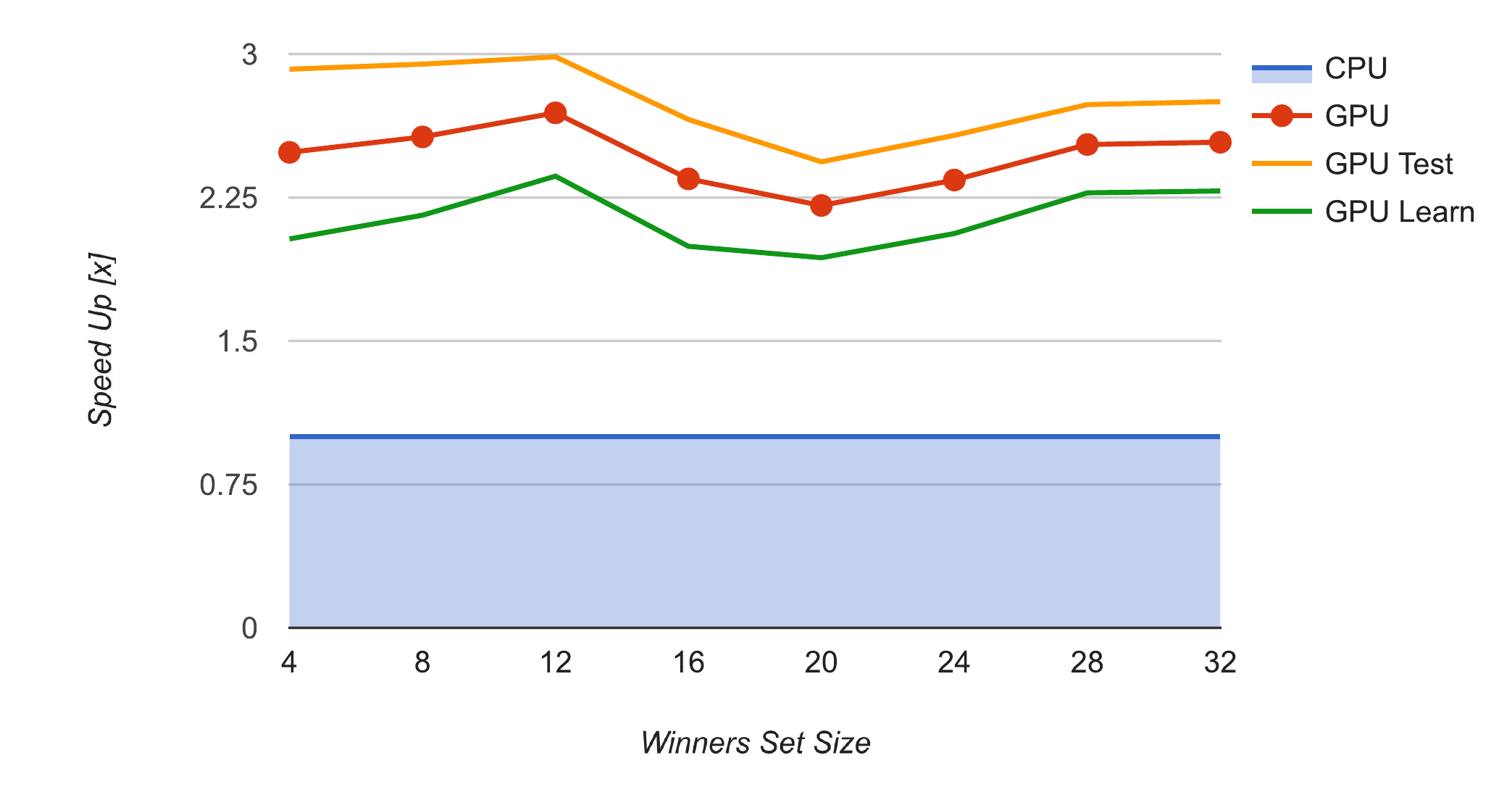}
    \caption{GPU vs CPU OCL}
\end{subfigure}
\begin{subfigure}{0.48\textwidth}
    \includegraphics[width=\textwidth]{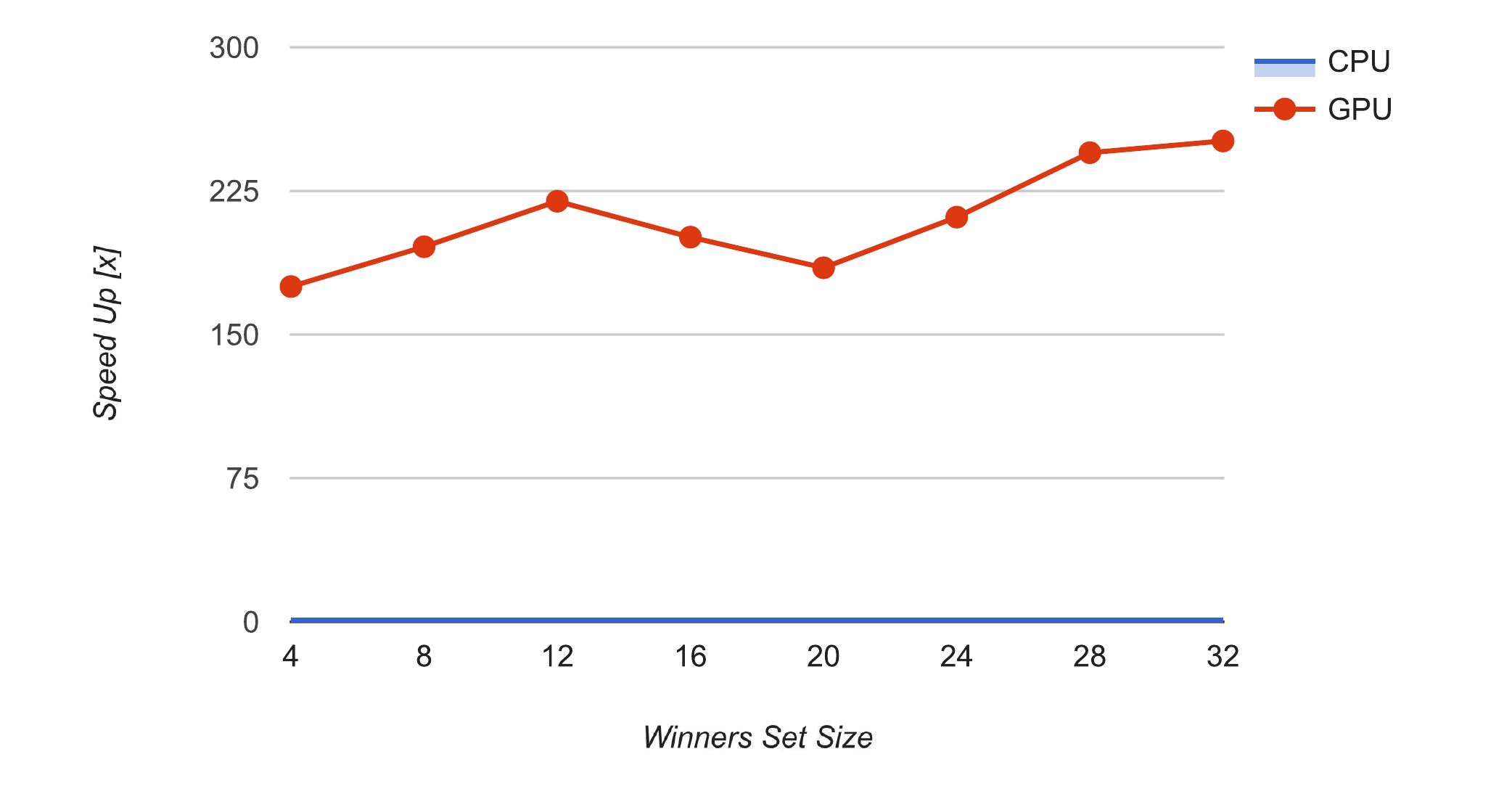}
    \caption{GPU vs CPU kernel}
\end{subfigure}
\begin{subfigure}{0.48\textwidth}
    \includegraphics[width=\textwidth]{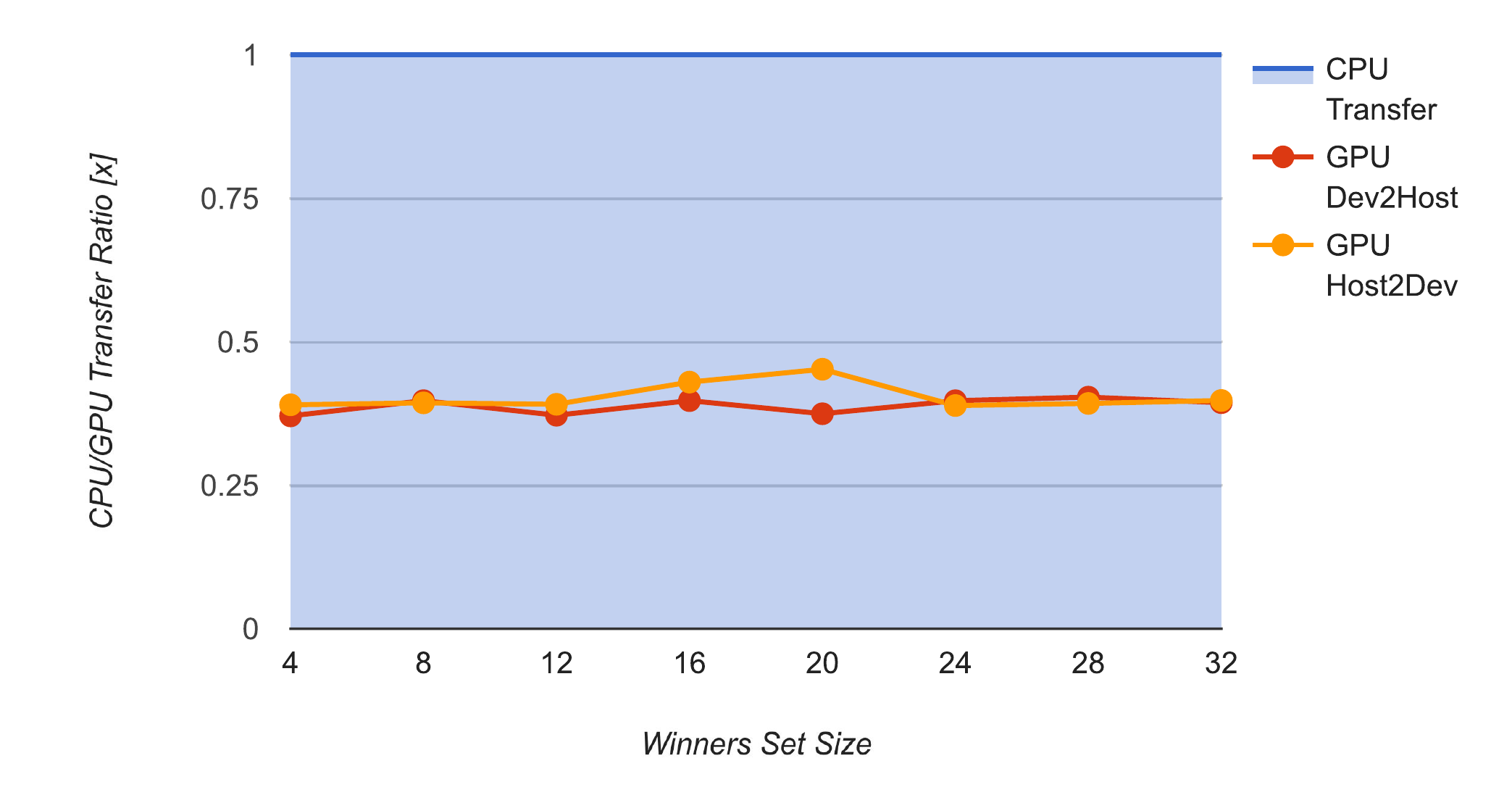}
    \caption{GPU vs CPU data transfer}
\end{subfigure}
\caption{Profiling results for winners set size}
\label{fig:pr_wss_speedup}
\end{figure}

It should be noted that the GPU supersedes OpenCL CPU inhibition implementation and the discrepancy increases with increasing column numbers as it was presented in Fig. \ref{fig:pr_columns_time}. Furthermore, OpenCL kernel performance is substantially better than its CPU counterpart (Fig. \ref{fig:pr_columns_speedup}). However, when kernel launching procedures and data transfer are taken into account the speed-up is reduced. It is worth noting that it levels off at about 130x and 2.5x for kernel and OCL tests, respectively.

Fig. \ref{fig:pr_synapses_time} and \ref{fig:pr_synapses_speedup} show a change of speed-up as a function of the number of synapses connected to each column of a Spatial Pooler. The more synapses are connected, the greater the acceleration that is achieved. This results from the internal architecture of the overlap module (Fig. \ref{fig:overlap}) which is, in essence, a hardware reduction operation performed within each GPU block. Fig. \ref{fig:pr_synapses_speedup} depicts that both learning and testing phases of SP yield the same speed-up results. It is worth noting that, depending on the accelerator, there is a constraint on a maximum size of a work group, which directly translates to a limit in the number of synapses that can be accommodated by a single GPU block.

\clearpage

$Min\_overlap$ has a slight impact on performance and speed--up of the object classification system (Fig. \ref{fig:pr_min_overlap_time} and \ref{fig:pr_min_overlap_speedup}). GPU execution time is gradually reduced reduced with a rise of $min\_overlap$. This results from the kernel implementation which allows for bypassing inhibition computation whenever overlap is lower than $min\_overlap$. For higher overlap values the number of zeros rapidly grows which leads to the rise of CPU/GPU speed-up.

$Winners\_set\_size$ is the number of 'winning' (having the highest overlap score) columns among the given column competitors in a contest to be chosen as active\cite{Numenta}. The number of neighboring columns which are taken into account impacts the computational effort since the columns are compared with all others within the inhibition range. Since $winners\_set\_size$ affects the inhibition radius, the larger the $winners\_set\_size$ is, the bigger the discrepancy in computation time between CPU and GPU, which is depicted in Fig. \ref{fig:pr_wss_speedup}. Winners set computation may be perceived as a specific kind of reduction operation.

\begin{figure}
\centering
\begin{subfigure}{0.48\textwidth}
    \includegraphics[width=\textwidth]{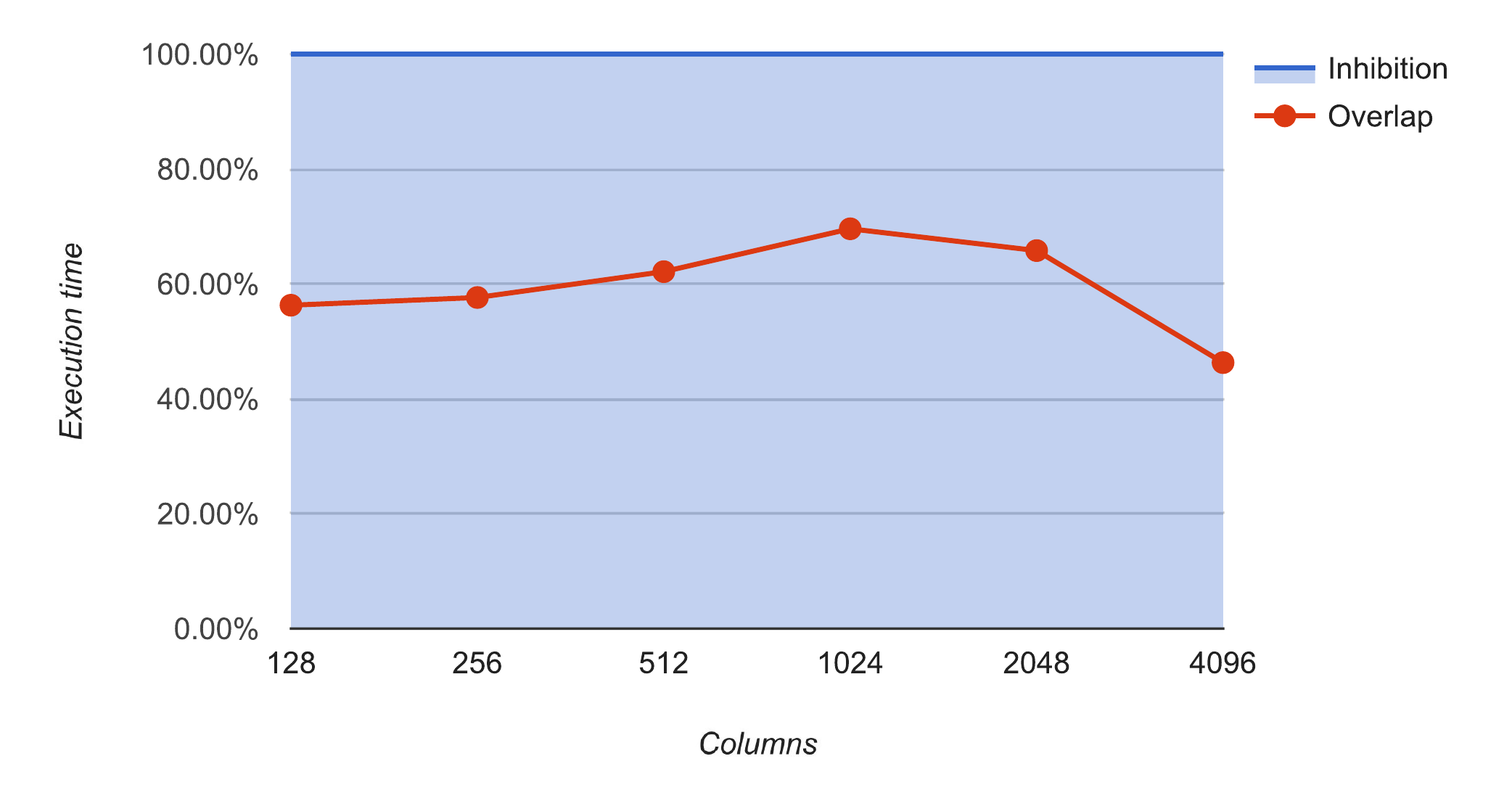}
\end{subfigure}
\begin{subfigure}{0.48\textwidth}
    \includegraphics[width=\textwidth]{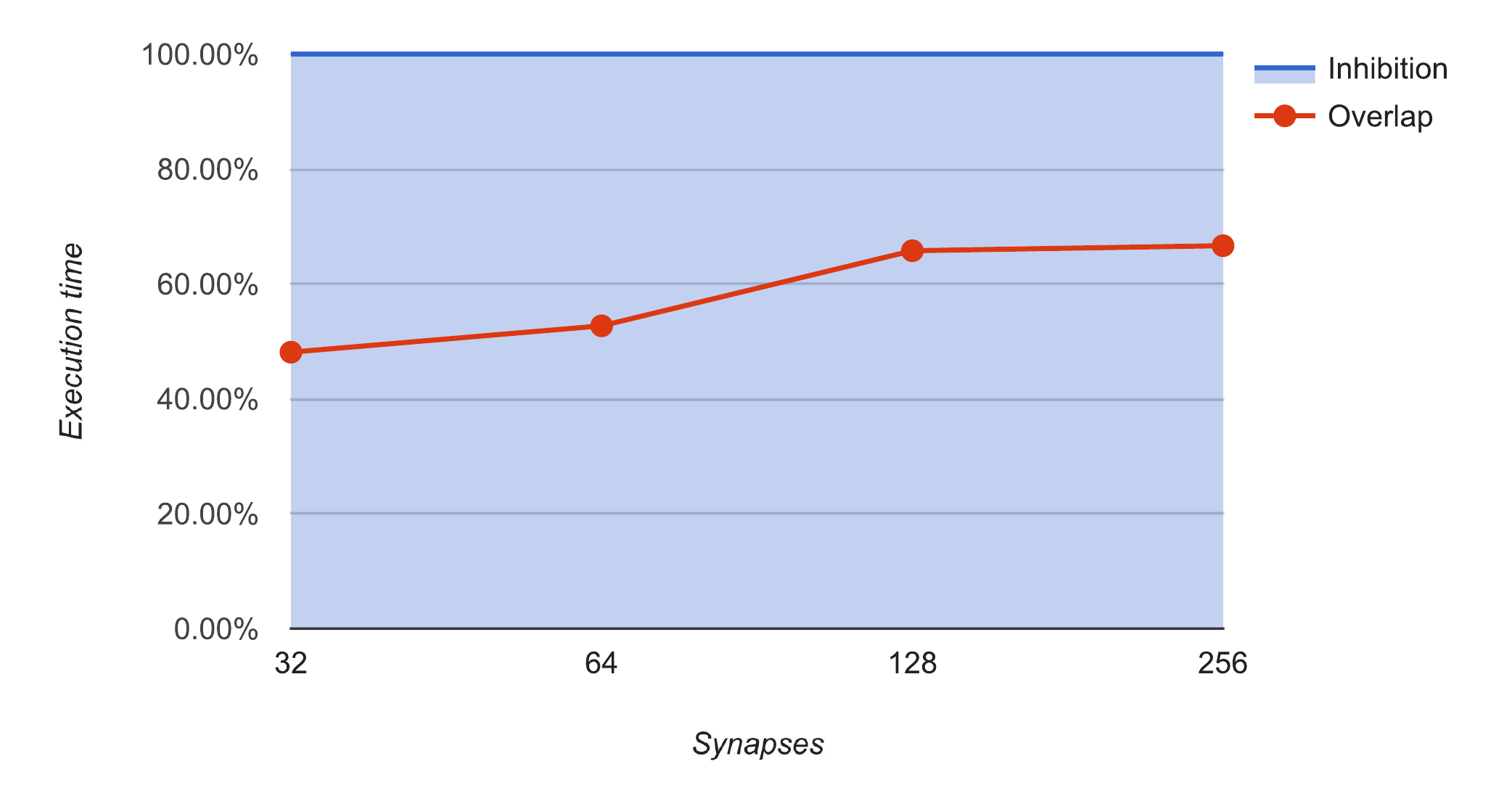}
\end{subfigure}
\caption{Percentage of Overlap kernel execution time in whole Inhibition kernel execution time (on GPU)}
\label{fig:pr_overlap_percentage}
\end{figure}

Fig. \ref{fig:pr_overlap_percentage} presents the contribution of overlap computations to the complete inhibition execution routine. It ranges between 50 \% and 75 \% of total inhibition kernel calculation time.

It is worth emphasizing that overall OCL test results depend on data transfer, which in turn is related to data representation. Therefore, changing from integer to boolean data type will result in approximately 32--fold reduction of the amount of data to be transferred to the accelerator. Such a transition is unfortunately not available for all the data which are sent to the device, for instance $boost$ is of a float type and can not be easily mapped to boolean.

According to the authors' knowledge, it is hard to find papers which directly correspond to the research conducted in this work. Nevertheless, we examined the following papers : \cite{Yue, KarpathyCVPR14, Zha} which present results of video classification using UCF-101 dataset. The best systems presented in those papers are based on various architectures of Convolutional Neural Networks (CNNs) and achieve accuracy of 80\% or more. It is worth emphasizing that despite similar performance in terms of the quality results, our test setup is different mostly in terms of the dataset used for the experiments. 
\section{Conclusions and future work}
\label{section:conclusions}
This paper presents experimental results of using an HTM--based system for object classification in video streams. The classification accuracy of the system was examined through a series of experiments and the performance was given in terms of an F1 score as a function of the number of columns, synapses, $min\_overlap$ and $winners\_set\_size$. The system achieves the highest F1-score of 0.95 and 0.91 for $min\_overlap=4$ and 256 synapses, respectively.  
We have also conduced a series of experiments with different hardware setups and measured CPU/GPU acceleration. The best kernel speed-up of 632x and 207x was reached for 256 synapses and 1024 columns. However, overall acceleration including transfer time was significantly lower and amounted to 6.5x and 3.2x for the same setup.

In future work, the authors are going to modify the preprocessing stage of the video processing flow and introduce TP. The authors are going to implement the most computationally--exhaustive routines in OpenCL and deploy the system on platforms equipped with GPU-- or FPGA--based acceleration. This will enable conduction of experiments using video with a lower image reduction ratio and larger datasets as well as stacking several layers of SP.

\section*{Acknowledgment}
I would like to thank my wife Urszula Wielgosz for her huge contribution to the preparation of the paper.

\bibliographystyle{ws-jcsc}
\bibliography{bibliography}

\begin{thebibliography}{10}

\bibitem{Mountcastle}
V.~Mountcastle, {The columnar organization of the neocortex}, {\em Brain} {\bf
  120}(apr 1997)  701--722.

\bibitem{humanbrainproject}
{The Human Brain Project - Human Brain Project}
  \url{https://www.humanbrainproject.eu}, (Accessed on 10.04.2016).

\bibitem{htm}
Custom {Hierarchical Temporal Memory} implementation
  \url{https://bitbucket.org/maciekwielgosz/htm-hardware-architecture},
  (Accessed on 12.04.2016).

\bibitem{Numenta}
J.~Hawkins, S.~Ahmad and D.~Dubinsky, {Hierarchical temporal memory including
  HTM cortical learning algorithms}, tech. rep., Numenta, Inc (sep 2011).

\bibitem{Chen}
X.~Chen, W.~Wang and W.~Li, {An overview of Hierarchical Temporal Memory: A new
  neocortex algorithm}, in {\em Modelling, Identification {\&} Control (ICMIC),
  2012 Proceedings of International Conference on\/},  IEEE, (Wuhan, China,
  2012), pp. 1004--1010.

\bibitem{Rachkovskij}
D.~Rachkovskij, {Representation and processing of structures with binary sparse
  distributed codes}, {\em IEEE Transactions on Knowledge and Data Engineering}
  {\bf 13}(2) (2001)  261--276.

\bibitem{Haibo}
H.~He and E.~A. Garcia, {Learning from imbalanced data}, {\em IEEE Transactions
  on Knowledge and Data Engineering} {\bf 21}(sep 2009)  1263--1284.

\bibitem{Peng}
P.~Zhang, X.~Zhu and L.~Guo, {Mining Data Streams with Labeled and Unlabeled
  Training Examples}, in {\em 2009 Ninth IEEE International Conference on Data
  Mining\/},  IEEE, (IEEE, Miami, USA, dec 2009), pp. 627--636.

\bibitem{Lu}
X.~Lu, C.~Zhang and X.~Yang, {Online video object classification using fast
  similarity network fusion}, in {\em 2014 IEEE Visual Communications and Image
  Processing Conference\/},  IEEE, (IEEE, Valletta, Malta, dec 2014), pp.
  346--349.

\bibitem{Hota}
R.~N. Hota, V.~Venkoparao and A.~Rajagopal, {Shape Based Object Classification
  for Automated Video Surveillance with Feature Selection}, in {\em 10th
  International Conference on Information Technology (ICIT 2007)\/},  IEEE,
  (IEEE, Rourkela, India, dec 2007), pp. 97--99.

\bibitem{Islam}
M.~K. Islam, F.~Jahan, J.-H. Min and J.-H. Baek, {Object classification based
  on visual and extended features for video surveillance application}, in {\em
  Control Conference (ASCC), 2011 8th Asian\/},  IEEE, (Kaohsiung, Taiwan,
  2011), pp. 1398--1401.

\bibitem{Castrill}
M.~Castrill{\'{o}}n, O.~D{\'{e}}niz, C.~Guerra and M.~Hern{\'{a}}ndez,
  {ENCARA2: Real-time detection of multiple faces at different resolutions in
  video streams}, {\em Journal of Visual Communication and Image
  Representation} {\bf 18}(apr 2007)  130--140.

\bibitem{Devarakota}
P.~Devarakota, M.~Castillo-Franco, R.~Ginhoux, B.~Mirbach, S.~Kater and
  B.~Ottersten, {3-D-Skeleton-Based Head Detection and Tracking Using Range
  Images}, {\em IEEE Transactions on Vehicular Technology} {\bf 58}(oct 2009)
  4064--4077.

\bibitem{Khan}
F.~N. Khan and S.~A. Khan, {Real-time object based single-stream to
  multi-stream network enabled multimedia system using an adder-less
  reconfigurable fast area correlator processor}, in {\em 8th International
  Multitopic Conference, 2004. Proceedings of INMIC 2004.\/},  IEEE, (IEEE,
  Lahore, Pakistan, 2004), pp. 688--693.

\bibitem{Bengio}
Y.~Bengio, A.~Courville and P.~Vincent, {Representation Learning: A Review and
  New Perspectives}, {\em IEEE Transactions on Pattern Analysis and Machine
  Intelligence} {\bf 35}(aug 2013)  1798--1828.

\bibitem{opencv_library}
G.~Bradski, {The OpenCV Library}, {\em Dr. Dobb's Journal of Software Tools}
  (2000).

\bibitem{blender}
{Blender project - Free and Open 3D Creation Software}
  \url{https://www.blender.org/}, (Accessed on 12.04.2016).

\bibitem{shapes_dataset}
{HTM Test Datasets} \url{http://data.wielgosz.info}, (Accessed on 02.07.2016).

\bibitem{kloeckner_pycuda_2012}
A.~{Kl{\"o}ckner}, N.~{Pinto}, Y.~{Lee}, B.~{Catanzaro}, P.~{Ivanov} and
  A.~{Fasih}, {PyCUDA and PyOpenCL: A Scripting-Based Approach to GPU Run-Time
  Code Generation}, {\em Parallel Computing} {\bf 38}(3) (2012)  157--174.

\bibitem{Yue}
{Joe Yue-Hei Ng}, M.~Hausknecht, S.~Vijayanarasimhan, O.~Vinyals, R.~Monga and
  G.~Toderici, {Beyond short snippets: Deep networks for video classification},
  {\em 2015 IEEE Conference on Computer Vision and Pattern Recognition (CVPR)}
  (jun 2015)  4694--4702.

\bibitem{KarpathyCVPR14}
A.~Karpathy, G.~Toderici, S.~Shetty, T.~Leung, R.~Sukthankar and L.~Fei-Fei,
  {Large-Scale Video Classification with Convolutional Neural Networks}, in
  {\em 2014 IEEE Conference on Computer Vision and Pattern Recognition\/},
  (IEEE, jun 2014), pp. 1725--1732.

\bibitem{Zha}
S.~Zha, F.~Luisier, W.~Andrews, N.~Srivastava and R.~Salakhutdinov, {Exploiting
  Image-trained CNN Architectures for Unconstrained Video Classification}, {\em
  ArXiv e-prints} (mar 2015).

\end{thebibliography}

\end{document}